\documentclass[runningheads]{llncs}
\usepackage{graphicx}
\usepackage{comment}
\usepackage{amsmath,amssymb} % define this before the line numbering.
\usepackage{color}
\usepackage{bbding}
\usepackage{pifont}
\usepackage{xcolor}
\usepackage{colortbl}
\usepackage{mathtools}
\usepackage{mathrsfs}
\usepackage{multirow}
\usepackage{algorithmic}
\usepackage{subfigure}
\usepackage{array}
\usepackage{tabu}
\usepackage{adjustbox}
\usepackage{tabularx}
\usepackage{booktabs}
\usepackage{diagbox}
\usepackage{array}
\usepackage{float}
 \usepackage{threeparttable}
\usepackage{makecell}
\usepackage[width=122mm,left=12mm,paperwidth=146mm,height=193mm,top=12mm,paperheight=217mm]{geometry}
\usepackage[pagebackref=true,breaklinks=true,letterpaper=true,colorlinks,bookmarks=false,urlcolor=blue]{hyperref}
\newcommand{\textBC}[2]{\textbf{\textcolor{#1}{#2}}}

\begin{document}
% \renewcommand\thelinenumber{\color[rgb]{0.2,0.5,0.8}\normalfont\sffamily\scriptsize\arabic{linenumber}\color[rgb]{0,0,0}}
% \renewcommand\makeLineNumber {\hss\thelinenumber\ \hspace{6mm} \rlap{\hskip\textwidth\ \hspace{6.5mm}\thelinenumber}}
% \linenumbers
\pagestyle{headings}
\mainmatter
\def\ECCVSubNumber{4160}  % Insert your submission number here

\title{A Single Stream Network for Robust and Real-time RGB-D Salient Object Detection} % Replace with your title

% INITIAL SUBMISSION 
\begin{comment}
\titlerunning{ECCV-20 submission ID \ECCVSubNumber} 
\authorrunning{ECCV-20 submission ID \ECCVSubNumber} 
\author{Anonymous ECCV submission}
\institute{Paper ID \ECCVSubNumber}
\end{comment}
%******************

% CAMERA READY SUBMISSION
%\begin{comment}
\titlerunning{A Single Stream Network for RGB-D Salient Object Detection}
% If the paper title is too long for the running head, you can set
% an abbreviated paper title here
%
\author{Xiaoqi Zhao\inst{1} \and
Lihe Zhang\inst{1}\protect\footnotemark[1]\and
Youwei Pang\inst{1}\and Huchuan Lu\inst{1,2}\and Lei Zhang\inst{3,4}}
\authorrunning{Zhao et al.}
% First names are abbreviated in the running head.
% If there are more than two authors, 'et al.' is used.
%
\institute{Dalian University of Technology, China 
\and
Peng Cheng Laboratory 
\and 
Dept. of Computing, The Hong Kong Polytechnic University, China
\and 
DAMO Academy, Alibaba Group \\
% \email{lncs@springer.com}\\
% \url{http://www.springer.com/gp/computer-science/lncs} \and
% ABC Institute, Rupert-Karls-University Heidelberg, Heidelberg, Germany\\
\email{\{zxq,lartpang\}@mail.dlut.edu.cn, \{zhanglihe,lhchuan\}@dlut.edu.cn,\ cslzhang@comppolyu.edu.hk}
\url{https://github.com/Xiaoqi-Zhao-DLUT/DANet-RGBD-Saliency}
}
%\end{comment}
%******************
\maketitle
\renewcommand{\thefootnote}{\fnsymbol{footnote}} %将脚注符号设置为fnsymbol类型，即特殊符号表示
% \footnotetext[2]{These authors contributed equally to this work.} %对应脚注[1]
\footnotetext[1]{Corresponding author.} %对应脚注[2]
\renewcommand{\thefootnote}{\arabic{footnote}}
\begin{abstract}
% \linespread{0.95}\selectfont
% \setlength{\parskip}{0.5\baselineskip}
Existing RGB-D salient object detection (SOD) approaches concentrate on the cross-modal fusion between the RGB stream and the depth stream. They do not deeply explore the effect of the depth map itself. In this work, we design a single stream network to directly use the depth map to guide early fusion and middle fusion between RGB and depth, which saves the feature encoder of the depth stream and achieves a lightweight and real-time model.        
We tactfully utilize depth information from two perspectives: (1) Overcoming the incompatibility problem caused by the great difference between modalities, we build a single stream encoder to achieve the early fusion, which can take full advantage of ImageNet pre-trained backbone model to extract rich and discriminative features. 
(2) We design a novel depth-enhanced dual attention module (DEDA) to efficiently provide the fore-/back-ground branches with the spatially filtered features, which enables the decoder to optimally perform the middle fusion. Besides, we put forward a pyramidally attended feature extraction module (PAFE) to accurately localize the objects of different scales. Extensive experiments demonstrate that the proposed model performs favorably against most state-of-the-art methods under different evaluation metrics. Furthermore, this model is 55.5\% lighter than the current lightest model and runs at a real-time speed of 32 FPS when processing a $384 \times 384$ image. 
\end{abstract}

\section{Introduction}
Salient object detection (SOD) aims to estimate  visual significance of image regions and then segment salient targets out. It has been widely used in many fields, \textit{e.g.}, scene classification~\cite{classification}, visual tracking~\cite{tracking}, person re-identification~\cite{Reid}, foreground maps evaluation~\cite{S-m}, content-aware image editing~\cite{image_editing}, light field image segmentation~\cite{LFSD_CNNs} and image captioning~\cite{Imagecaption}, etc.
\begin{figure}[t]
\centering
\includegraphics[width=0.5\linewidth,height=0.3\linewidth]{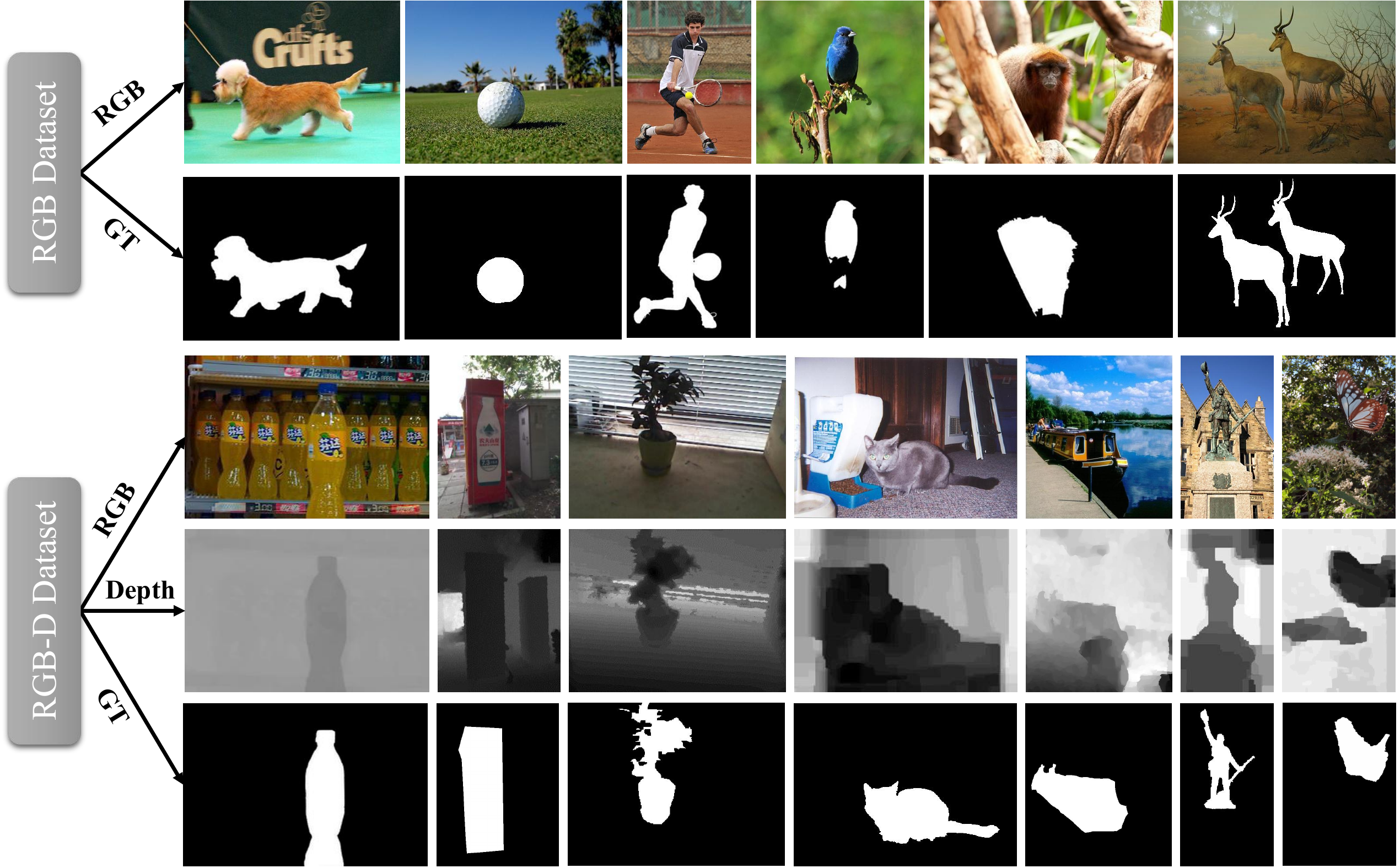}\\
        %\vspace{-3mm}
\caption{Visual comparison of RGB and RGB-D SOD datasets.}
\label{fig:RGB_RGBD_Datasets}
%\vspace{-4mm}
% \vspace{-5.5mm}
\end{figure}

With the development of deep convolutional neural networks (CNNs), a large number of CNN-based methods~\cite{RFCN,Amulet,SRM,DGRL,BMPM,RAS,PAGRN,HRS,PASE,BASNet,CTBIN,MINet,GateNet} have been proposed for RGB salient object detection and they achieve satisfactory performance. However,  some complex scenarios are still unresolved, such as salient objects share similar appearances to the background or the contrast among different objects is extremely low. Under these circumstances, only using the information provided by the RGB image is not sufficient to predict saliency map well. Recently, benefiting from Microsoft Kinect and Intel RealSense devices, depth information can be conveniently obtained. Moreover, the stable geometric structures depicted in the depth map are robust against the changes of illumination and texture, which can provide important supplement information for handling complex environments, as shown in Fig.~\ref{fig:RGB_RGBD_Datasets}. These examples in the RGB-D dataset have more stereoscopic viewing angles and more severe interference from the background than ones in the RGB dataset. 

For the RGB-D SOD task, many CNN-based methods~\cite{PCA,MMCI,TANet,CPFP,DMRA,HDFNet} are proposed, but more efforts need be paid to achieve a robust, real-time and small-scale model. We analyze their restrictions here: (1) Most methods~\cite{CTMF,PDNet,PCA,AF_RGBD,MMCI,DMRA} use the two-stream structure to separately extract features from RGB and depth, which greatly increases the number of parameters in the network. 
In addition, due to small scale of existing RGB-D datasets and great difference between RGB and depth modalities, the deep network (e.g., VGG, ResNet) is very difficult to be trained from scratch if the RGB and depth channels are concatenated and fed into the network.
To this end, we construct a single stream encoder, which can borrow the generalization ability of ImageNet pre-trained backbone to extract discriminative features from the RGB-D input and achieve SOD-oriented RGB-depth early fusion.    
(2)
The depth map can naturally depict contrast cues at different positions, which provides important guidance for the fore-/back-ground segmentation. However, this observation has never been investigated in the existing literature. In this work,
we introduce a spatial filtering mechanism between the encoder and the decoder, which explicitly utilizes the depth map to guide the computation of dual attention, thereby promoting feature discrimination in the fore-/back-ground decoding branches. 
(3) Since the size of objects is various, %an image usually contains multi-scale objects
the effective utilization of multi-scale contextual information is very key to accurately localize objects. Previous methods~\cite{SRM,R3Net,BMPM,PFA,DMRA} 
do not explore the internal relationships between the parallel features of different receptive fields in the multi-scale feature extraction module (e.g. ASPP~\cite{Deeplab}). 
We think that each position in the feature map responds differently to objects and a strong activation area can better perceive the semantic cues of objects.

To address these above problems, we propose a single stream network with the novel depth-enhanced  attention (DANet) for RGB-D saliency detection. First, we design a single stream encoder with a 4-channel input. It can not only save many parameters compared to previous two-stream methods, but also promote the regional discrimination of the low-level features because this encoder can effectively utilize the ImageNet pre-trained model to extract powerful features with the help of the proposed initialization strategy. Second, we build a depth-enhanced dual attention module (DEDA) between the encoder and the decoder. 
This module sequentially leverages both the mask-guided strategy and the depth-guided strategy to filter the mutual interference between depth prior and appearance prior, thereby enhancing the overall contrast between  foreground and background.   
In addition, we present a pyramidal attention mechanism to promote the representation ability of the top-layer features. It calculates the spatial correlation among different scales and obtains efficient context guidance for the decoder. 

Our main contributions are summarized as follows.
\begin{itemize}
     \item We propose a single stream network  to achieve both early fusion and middle fusion, which implicitly formulates the cross-modal information interaction in the encoder and further explicitly enhances this effect in the decoder. 
     
    \item We design a novel depth-enhanced dual attention mechanism, which exploits the depth map to  strengthen the mask-guided attention and computes fore-/back-ground attended features for the encoder.
    
    \item Through using a self-attention mechanism, we propose a  pyramidally attended feature extraction module, which can depict spatial dependencies between any two positions in feature map.
    
    \item We compare the proposed model with ten state-of-the-art RGB-D SOD methods on six challenging datasets. The results show that our method performs much better than other competitors. Meanwhile, the proposed model is much lighter than others and achieves a real-time speed of 32 FPS.    
\end{itemize}

\section{Related Work}
Generally speaking, the depth map can be utilized in three ways: early fusion~\cite{early_fusion_1,early_fusion_2}, middle fusion~\cite{Middle_fusion} and late fusion~\cite{late_fusion}. It is worth noting that the early fusion technique has not been explored in existing deep learning based saliency methods.  Most of them use two streams to respectively handle RGB and depth information. They achieve the cross-modal fusion only at a specific stage, which limits the usage of the depth-related prior knowledge. This issue motivates some efforts~\cite{PCA,MMCI} to examine the multi-level fusion between the two streams. 
However, the two-stream design significantly increases the number of parameters in the network~\cite{CTMF,PCA,MMCI,AF_RGBD}. And, restricted by the scale of existing RGB-D datasets, the depth stream is hardly effectively trained and does not comprehensively capture depth cues to guide salient object detection. 
To this end, Zhao \textit{et al.}~\cite{CPFP} propose a trade-off method, which only feeds the RGB images into the encoder network and inserts a shallow convolutional subnet between adjacent encoder blocks to extract the guidance information from the depth map. 
In this work, we integrate the depth map and the RGB image from starting to build a real single-stream network. This network can fully use the advantage of the ImageNet pre-trained model to extract color and depth features and remedy the deficiencies of individual grouping cues in color space and depth space. And we also show the effectiveness of the proposed early fusion strategy in the encoder through quantitative and qualitative analysis.
Recently, Zhao \textit{et al.}~\cite{CPFP} exploit the depth map to compute a contrast prior and then use this prior to enhance the encoder features. Their contrast loss actually enforces the network to learn saliency cues from the depth map in a brute-force manner. 
Although the resulted attention map can coarsely distinguish the foreground from the background, it greatly reduces the ability of providing accurate depth prior for some easily-confused regions, thereby weakening the discrimination of the encoder feature in these regions.
We think that the depth map is more suitable to play a guiding role because the grouping cues in depth space are very incompatible with those in color space.
In this work, we combine the depth guidance and the mask guidance to explicitly formulate their complementary relation. Thus, we can effectively take advantage of the useful depth cues to assist in segmenting salient objects and weaken their incompatibility.   

\begin{figure*}
    \includegraphics[width=0.8\textwidth, height=0.4\linewidth]{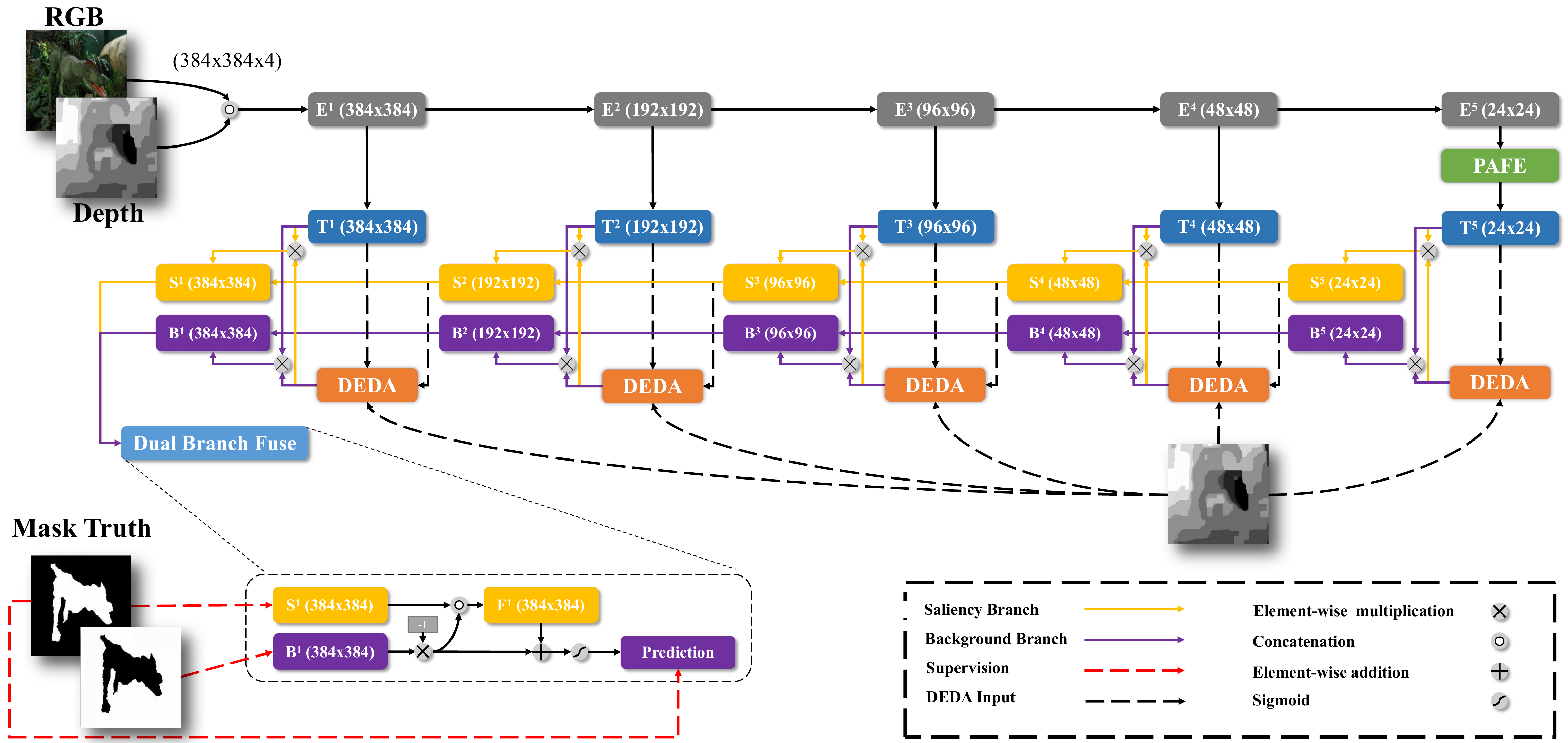}\\ %,height=0.3\linewidth
    %      \vspace{-8mm}
    \centering
    \caption{Network pipeline. It consists of the VGG-16 ($\mathbf{E}^1 \sim \mathbf{E}^5$), five transition layers ($\mathbf{T}^1 \sim \mathbf{T}^5$), five saliency layers ($\mathbf{S}^1 \sim \mathbf{S}^5$), five background layers ($\mathbf{B}^1 \sim \mathbf{B}^5$), the pyramidally attended feature extraction module (PAFE) and  the depth-enhanced dual attention module (DEDA). The final prediction is generated by using residual connections to fuse the outputs from  $\mathbf{S}^1$ and $\mathbf{B}^1$.} 		
    \label{fig:DANet}
\end{figure*} 

\section{Proposed Method}
We adopt the feature pyramid network~\cite{FPN} (FPN) as the basic structure and the overall architecture is shown in Fig.~\ref{fig:DANet}, in which encoder blocks, transition layers, saliency layers and background layers are denoted as $\mathbf{E}^i$, $\mathbf{T}^i$, $\mathbf{S}^i$ and $\mathbf{B}^i$, respectively. Here, $i \in \left \{1, 2, 3, 4, 5 \right \}$ indexes different levels. And their output feature maps are denoted as $E^i$, $T^i$, $S^i$ and $B^i$, respectively. Each transition layer uses a $3 \times 3$ convolution operation to process the features maps from each encoder block for matching the number of channels. The saliency layers and background layers compose the decoder. The final output is generated by integrating the predictions of the two branches using a residual connection. In this section, we first describe the encoder network in Sec.~\ref{sec:Encoder_Network}, then give the details of the proposed modules, including depth-enhanced dual attention module (DEDA) in Sec.~\ref{sec:DAM} and pyramidally attended feature extraction module (PAFE)  in Sec.~\ref{sec:PAFEM}.  
\subsection{Single Stream Encoder Network}\label{sec:Encoder_Network}
In our model, the encoder is a single stream with a FCN structure. We take the VGG-16~\cite{VGG} network as the backbone, which contains $13$ Conv layers, $5$ max-pooling layers and $2$ fully connected layers. First, we concatenate the depth map with the RGB image as the 4-channel RGB-D input. 
We initialize the parameters of the first convolutional layer in block $\mathbf{E}^1$ using the He's method~\cite{PRelu} and output a 64-channel feature. The other layers adopt the ImageNet pre-trained parameters. In this way, the two-modality information can be fused in the input stage and make the low-level features have a more powerful discriminant ability, which is conducive to extracting effective features for salient regions. 
Moreover, because four input channels are parallel in the channel direction, the network can easily learn to suppress the feature response of the depth channel when the quality of the depth map is poor and does not affect feature computation of the color channels. To demonstrate the effectiveness of this design, we compare two other schemes. Both of them combine the color channels with the depth channel by element-wise addition. One is to directly load the pre-trained parameters. The other is to use the above-mentioned parameter setting. When the depth map has a negative impact, the first layer simultaneously suppresses the color response and the depth response. The quantitative results in Tab.~\ref{tab:ablation} show that our early fusion strategy performs better than other schemes.  
Similar to most previous methods~\cite{PDNet,PCA,CPFP,TANet,AF_RGBD,DMRA}, we cast away all the fully-connected layers of the VGG-16 net and remove the last pooling layer to retain the details of the top-layer features.

\begin{figure}[t]
\centering
\includegraphics[width=0.55\linewidth,height=0.45\linewidth]{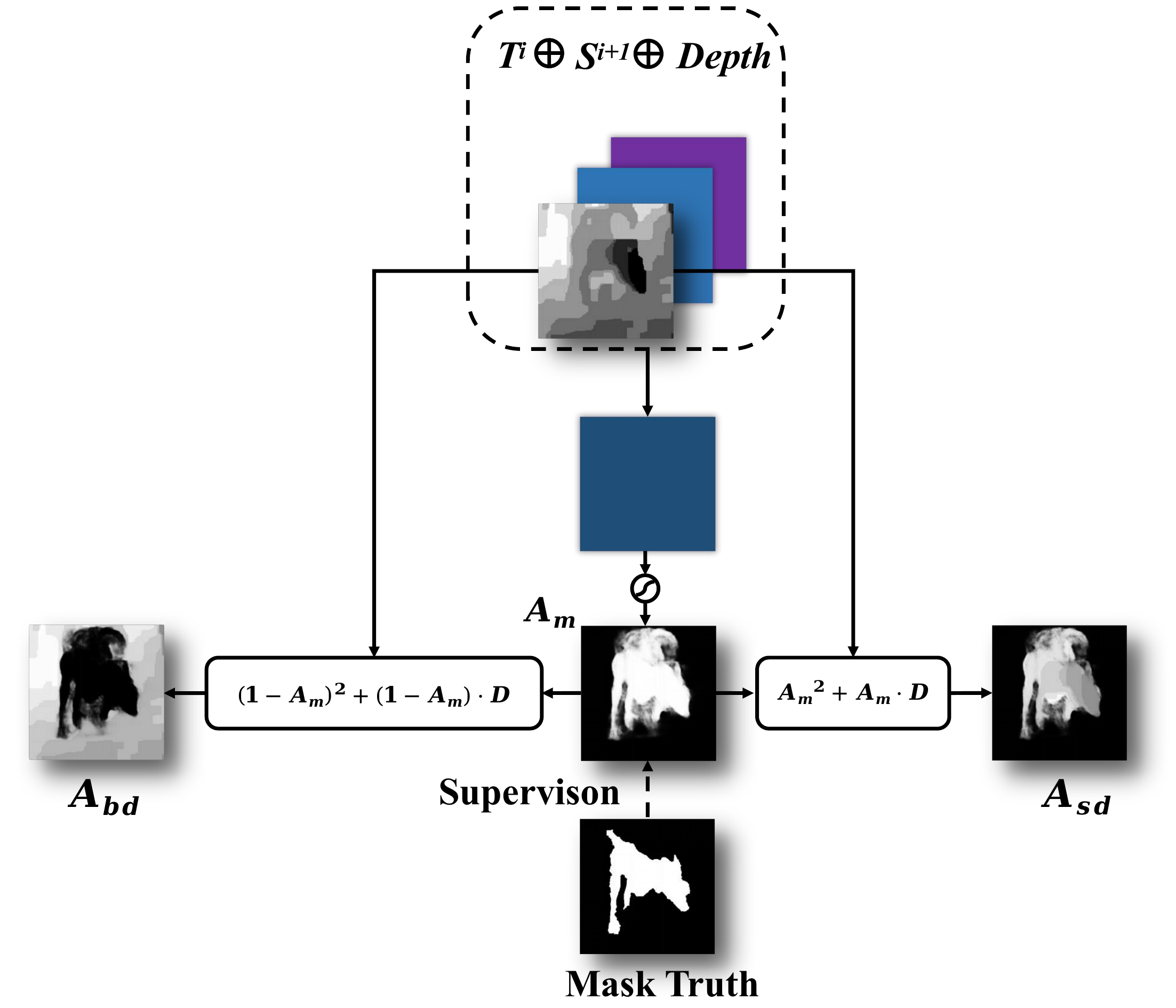}\\
\caption{Detailed diagram of depth-enhanced dual attention module.}
\label{fig:DAM}
\end{figure}
\subsection{Depth-enhanced Dual Attention Module}\label{sec:DAM}
Considering that the depth map can naturally describe contrast information in different depth positions, we utilize it to generate contrasted features for the decoder, thereby strengthening the detection ability for hard examples. In particular, we propose a depth-enhanced attention module and its detailed structure is shown in Fig.~\ref{fig:DAM}.
When the region of object has a large span at depth or the background and foreground areas are at the same depth, only depending on the depth map does not provide accurate grouping cues for saliency detection. Therefore, we adopt the mask supervision and depth guidance mechanism to filter the misleading information. 
We first combine the features from the current transition layer and the previous decoder block with the depth map to compute a mask-guided attention $A_{m}$, which is supervised by the saliency ground truth. The whole process is written as follows:
\begin{equation}\label{equ:1}
\centering
    A_{m}
    = \left\{\begin{matrix}
    \delta(Conv(T^{i}+S^{i+1}+D)) & \text{ if } i=1, 2, 3, 4\\
    \delta(Conv(T^{i}+D)) & \text{ if } i=5,
    \end{matrix}\right.
\end{equation}
where $\delta(\cdot)$ is an element-wise sigmoid function, $Conv(\cdot)$ refers to the convolution layer and $D$ denotes the depth map.  
Although the resulted $A_{m}$ shows high contrast between the foreground and the background under binary supervision, it inevitably exists two drawbacks: (1) Some background regions are wrongly classified to be salient. (2) Some salient regions are mislabelled as the background.
To solve the first issue, we introduce the depth information to refine $A_{m}$:
\begin{equation}\label{equ:2}
\begin{split}
A_{sd} = A_{m} \cdot A_{m}+ A_{m}\cdot D, \\
\end{split} 
\end{equation}
where $A_{sd}$ denotes the depth-enhanced attention of the saliency branch. 
It can provide additional contrast guidance for the misjudged regions in $A_{m}$ and maintain high contrast between foreground and background, thereby enhancing mask-guided attention.
To resolve the second issue, we design the depth-enhanced attention $A_{bd}$ for the background branch as follows:
\begin{equation}\label{equ:3}
\begin{split}
A_{bd} = (1-A_{m}) \cdot (1- A_{m})+(1- A_{m}) \cdot D.
\end{split} 
\end{equation}

We combine $A_{m}$ and $D$ by the above formulas to construct foreground attention $A_{sd}$ and background attention $A_{bd}$. 
There are three benefits: (1) When the depth value is very small or even zero, the attention still work because the first terms in Equ. (\ref{equ:2}) and Equ. (\ref{equ:3}) are independent of $D$. (2) The depth map does not have the semantic distinction between foreground and background, which may introduce noise and interference when segmenting salient object. However, the DEDA can still preserve high contrast between the foreground and the background while introducing depth information in Equ. (\ref{equ:2}) and Equ. (\ref{equ:3}). Becasue, the $A_{m}$ usually shows high contrast between the foreground and the background under binary supervision. $A_{m}$ $\cdot$ $D$ or $1-A_{m}$ $\cdot$ $D$ can limit $D$ to only optimize the foreground or the background.
(3) During the back-propagation process of gradient, $A_{sd}$ and $A_{bd}$ can obtain dynamic gradients, which help the network learn the optimal parameters. Taking  $A_{sd}$ for example, its derivation with respect to $A_{m}$ is calculated as:
\begin{equation}\label{equ:4}
\begin{split}
\frac{\mathrm{d}A_{sd}}{\mathrm{d}A_{m}} = 2 \cdot A_{m} + D,
\end{split} 
\end{equation}
from where it can be seen that the gradient changes with $A_{m}$ although the depth $D$ is fixed. 

\begin{figure}[t]
\centering
\includegraphics[width=0.58\linewidth,height=0.40\linewidth]{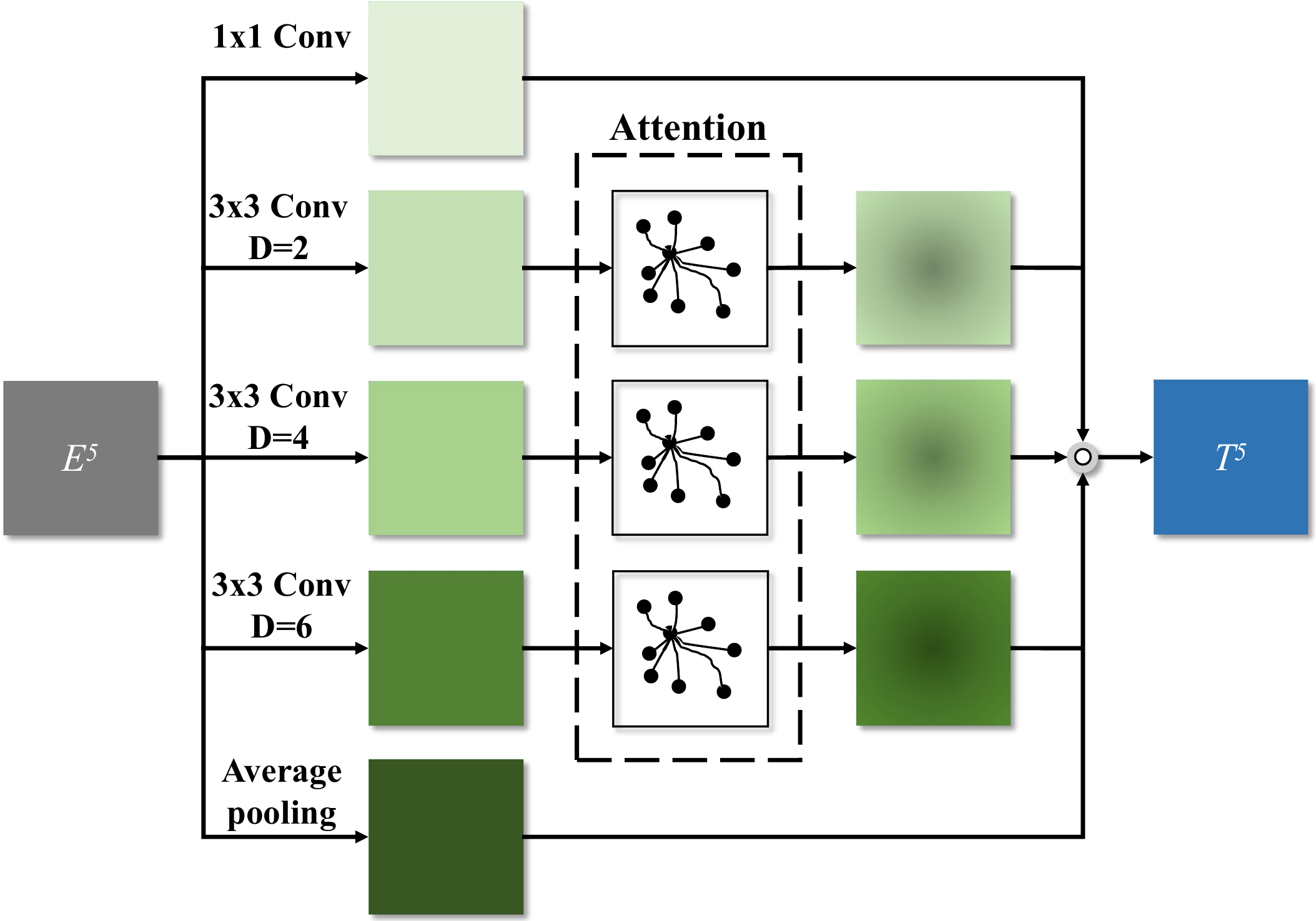}\\

\caption{Illustration of pyramidally attended feature extraction.}
\label{fig:PAFEM}
\end{figure}

\subsection{Pyramidally Attended Feature Extraction}\label{sec:PAFEM}
The scale of objects is various in images. The single-scale features can not capture the multi-scale context for different objects. Benefiting from the ASPP in semantic segmentation~\cite{Deeplab}, some SOD networks~\cite{R3Net,BMPM,PFA} also equip it. 
However, directly aggregating features at different scales may weaken the representation ability for salient areas because of the distraction of non-salient regions. Instead of equally treating all spatial positions, we respectively apply spatial attention to the features of different scales in order to focus more on the visually important regions.
By integrating the attention-enhanced multi-scale features, we build a pyramidally attended feature extraction module (PAFE). Its detailed structure is shown in Fig.~\ref{fig:PAFEM}.

We first load in parallel several dilated convolutional layers with different dilation rates on the top-layer $\mathbf{E}^5$ to extract high-level and multi-scale features. 
Then, an attention module is followed in individual branch. Our attention design is inspired by the non-local idea~\cite{Nonlocal}. We consider the pairwise relationship at any point in feature map to calculate the attention weight. Let $F_{in}  \in \mathbb{R}^{C \times H \times W}$ and $F_{out}  \in \mathbb{R}^{C \times H \times W}$ represent the input and the output of the attention module, respectively.  The attention map $A$ is computed as follows:
\begin{equation}\label{equ:5}
\begin{split}
A  = softmax(& R_1(Conv(F_{in}))^\top \\
  \times & R_1(Conv(F_{in}))), 
\end{split} 
\end{equation}
where $softmax(\cdot)$ is an element-wise softmax function and $R_1(\cdot)$ reshapes the input feature to $\mathbb{R}^{C \times N}$. $N = {H \times W}$ is the number of features.  

Next, we combine $A$  with  $F_{in}$ to yield the attention-enhanced feature map and then add the input $F_{in}$ to obtain the output $F_{out}$ as follows: 
\begin{equation}\label{equ:6}
\begin{split}
F_{out} = F_{in}+R_2(R_1(Conv(F_{in})) \times A^\top),
\end{split} 
\end{equation}
where $R_2(\cdot)$ reshapes the input feature to $\mathbb{R}^{C \times H \times W}$.  
In particular, the $1 \times 1$ convolution branch and the global average pooling branch aim to maintaining the inherent properties of the input by respectively using the minimal and maximum receptive field. Therefore, we do not apply the attention module to the two branches.

\section{Experiments}
\subsection{Dataset}
 We evaluate the proposed model on six public RGB-D SOD datasets which are \textbf{NJUD}~\cite{NJU2000}, \textbf{RGBD135}~\cite{RGBD135} \textbf{NLPR}~\cite{early_fusion_1}, \textbf{SSD}~\cite{SSD}, \textbf{DUTLF-D}~\cite{DMRA} and \textbf{SIP}~\cite{SIP}.
On the DUTLF-D, we adopt the same way as the DMRA~\cite{DMRA} to use $800$ images for training and the rest $400$ for testing. Following most state-of-the-art methods~\cite{PCA,MMCI,CTMF,CPFP}, we randomly select $1400$ samples from the NJUD dataset and $650$ samples from the NLPR dataset for training. Their remaining images and other three datasets are used for testing. 
\subsection{Evaluation Metrics} We adopt several widely used metrics for quantitative evaluation: precision-recall (PR) curves, F-measure score, mean absolute error (MAE, $\mathcal{M}$), the recently released S-measure ($S_{m}$) and E-measure ($E_{m}$) scores. The lower value is better for the MAE and higher is better for others.
\textbf{Precision-Recall curve}: The pairs of precision and recall are calculated by comparing the binary saliency maps with the ground truth to plot the PR curve, where the threshold for binarizing slides from $0$ to $255$.
\textbf{F-measure}: It is a metric that comprehensively considers both precision and recall:
 \begin{equation}\label{equ:7}
 \text{F}{_\beta } = \frac{{\left( {1 + {\beta ^2}} \right) \cdot \text{precision} \cdot \text{recall}}}{{{\beta ^2} \cdot \text{precision} + \text{recall}}},
 \end{equation}
  where $\beta^2$ is set to 0.3 as suggested in~\cite{colorcontrast_Fm} to emphasize the precision. In this paper, we report the maximum F-measure ($F_{\beta}^{max}$) score across the binary maps of different thresholds, the mean F-measure ($F_{\beta}^{mean}$) socre across an adaptive threshold and the weighted F-measure ($F_{\beta}^{w}$)~\cite{Fwb}.
\textbf{Mean Absolute Error}: It is a complement to the PR curve and measures the average absolute difference between the prediction and the  ground truth pixel by pixel.
\textbf{S-measure}: It evaluates the spatial structure similarity by combining the region-aware structural similarity  $S _ { r }$ and the object-aware structural similarity  $S _ { o }$: 
\begin{equation}\label{equ:8}
\text{S}{_m} = \alpha * S _ { o } + ( 1 - \alpha ) * S _ { r },
\end{equation}
where $\alpha$ is set to 0.5~\cite{S-m}.
\textbf{E-measure}: The enhanced alignment measure~\cite{Em} can jointly capture image level statistics and local pixel matching information.
\\
\subsection{Implementation Details} 
 Our model is implemented based on the Pytorch toolbox and trained on a PC with GTX 1080Ti GPU for $40$ epochs with mini-batch size $4$. The input RGB image and depth map are both resized to $384 \times 384$. For the RGB image, we use some data augmentation techniques to avoid overfitting: random horizontally flip, random rotate, random brightness, saturation and contrast. For the optimizer, we adopt the stochastic gradient descent (SGD) with a momentum of $0.9$ and a weight decay of $0.0005$. The learning rate is set to $0.001$ and later use the ``poly'' policy~\cite{poly} with the power of $0.9$ as a mean of adjustment. In this paper, we  use the binary cross-entropy loss as supervision.
\subsection{Comparison with State-of-the-art Results}
The performance of the proposed model is compared
with ten state-of-the-art approaches on six benchmark datasets, including the DES~\cite{RGBD135}, DCMC~\cite{DCMC}, CDCP~\cite{CDCP}, DF~\cite{DF}, CTMF~\cite{CTMF}, PCA~\cite{PCA}, MMCI~\cite{MMCI}, TANet~\cite{TANet}, CPFP~\cite{CPFP} and DMRA~\cite{DMRA}. For fair comparisons, all the saliency maps of these methods are directly provided by authors or computed by their released codes.
\begin{table*}[ht]
\large
  \caption{
  Quantitative comparison. $\uparrow$ and $ \downarrow$ indicate that the larger and smaller scores are better, respectively. Among the CNN-based methods, the best results are shown in $\textBC{red}{red}$. The subscript in each model name is the publication year.
  }
  \label{tab:scores}
   \renewcommand\tabcolsep{5.0pt} % 调整表格列间的宽度
   \renewcommand\arraystretch{1.5}
  \centering

\resizebox{0.92\textwidth}{!}  
   %\scalebox{.84}
  {
\begin{tabular}{ll|lll|lllllll|ll}

   \toprule[2pt]
 
\multicolumn{2}{l|}{\multirow{2}{*}{Metric}} & \multicolumn{3}{c|}{\textbf{\footnotesize{Traditional Methods}}} & \multicolumn{7}{c|}{\textbf{\footnotesize{VGG-16}}} & \multicolumn{2}{c}{\textbf{\footnotesize{VGG-19}}} \\
\multicolumn{2}{l|}{}   & \Large{DES$_{14}$}         & \Large{DCMC$_{16}$}         &\Large{ CDCP$_{17}$}     &\Large{DF$_{17}$} &\Large{ CTMF$_{18}$} & \Large{PCANet$_{18}$} & \Large{MMCI$_{19}$} &\Large{TANet$_{19}$} & \Large{CPFP$_{19}$} & \Large{DANet} &\Large{ DMRA$_{19}$} &\Large{DANet}   \\
\multicolumn{2}{l|}{}   
&  \multicolumn{1}{c}{\Large{~\cite{RGBD135}}}         &    \multicolumn{1}{c}{\Large{~\cite{DCMC}}}           &   \multicolumn{1}{c|}{\Large{~\cite{CDCP}}}           &  \multicolumn{1}{c}{\Large{~\cite{DF}}}   & \multicolumn{1}{c}{\Large{~\cite{CTMF}}}      &  \multicolumn{1}{c}{\Large{~\cite{PCA}}}       &   \multicolumn{1}{c}{\Large{~\cite{MMCI}}}    &   \multicolumn{1}{c}{\Large{~\cite{TANet}}}     &   \multicolumn{1}{c}{\Large{~\cite{CPFP}}}    &   \multicolumn{1}{c|}{Ours}    &   \multicolumn{1}{c}{\LARGE{~\cite{DMRA}}}       &  \multicolumn{1}{c}{Ours}  \\
\hline
\multirow{6}{*}{\emph{\rotatebox{90}{SSD~\cite{SSD}}}}      
&$F_{\beta}^{max}\uparrow$ & \multicolumn{1}{c}{\Large{0.260}} &  \multicolumn{1}{c}{\Large{0.750}}    & \multicolumn{1}{c|}{\Large{0.576}}   &  \multicolumn{1}{c}{\Large{0.763}}   &   \multicolumn{1}{c}{\Large{0.755}}    & \multicolumn{1}{c}{\Large{0.844}}  &\multicolumn{1}{c}{\Large{0.823}}  &  \multicolumn{1}{c}{\Large{0.835}}      &  \multicolumn{1}{c}{\Large{0.801}}     &    \multicolumn{1}{c|}{\textBC{red}{\Large{0.888}}}   &  \multicolumn{1}{c}{\Large{0.858}}     &   \multicolumn{1}{c}{\textBC{red}{\Large{0.866}}}     \\
&$F_{\beta}^{mean}\uparrow$ & \multicolumn{1}{c}{\Large{0.073}} &  \multicolumn{1}{c}{\Large{0.684}}    & \multicolumn{1}{c|}{\Large{0.524}}   &  \multicolumn{1}{c}{\Large{0.709}}   &   \multicolumn{1}{c}{\Large{0.709}}    & \multicolumn{1}{c}{\Large{0.786}}  &\multicolumn{1}{c}{\Large{0.748}}  &  \multicolumn{1}{c}{\Large{0.767}}      &  \multicolumn{1}{c}{\Large{0.726}}     &    \multicolumn{1}{c|}{\textBC{red}{\Large{0.831}}}   &  \multicolumn{1}{c}{\Large{0.821}}     &   \multicolumn{1}{c}{\textBC{red}{\Large{0.827}}}    \\
&$F_{\beta}^{w}\uparrow$   & \multicolumn{1}{c}{\Large{0.172}} &  \multicolumn{1}{c}{\Large{0.480}}    & \multicolumn{1}{c|}{\Large{0.429}}   &  \multicolumn{1}{c}{\Large{0.536}}   &   \multicolumn{1}{c}{\Large{0.622}}    & \multicolumn{1}{c}{\Large{0.733}}  &\multicolumn{1}{c}{\Large{0.662}}  &  \multicolumn{1}{c}{\Large{0.727}}      &  \multicolumn{1}{c}{\Large{0.709}}     &    \multicolumn{1}{c|}{\textBC{red}{\Large{0.798}}}   &  \multicolumn{1}{c}{\Large{0.787}}     &   \multicolumn{1}{c}{\textBC{red}{\Large{0.795}}}      \\
& $S_m\uparrow$        & \multicolumn{1}{c}{\Large{0.341}} &  \multicolumn{1}{c}{\Large{0.706}}    & \multicolumn{1}{c|}{\Large{0.603}}   &  \multicolumn{1}{c}{\Large{0.741}}   &   \multicolumn{1}{c}{\Large{0.776}}    & \multicolumn{1}{c}{\Large{0.842}} &\multicolumn{1}{c}{\Large{0.813}}  &  \multicolumn{1}{c}{\Large{0.839}}      &  \multicolumn{1}{c}{\Large{0.807}}     &    \multicolumn{1}{c|}{\textBC{red}{\Large{0.869}}}   &  \multicolumn{1}{c}{\Large{0.856}}     &   \multicolumn{1}{c}{\textBC{red}{\Large{0.864}}}   \\
& $E_m\uparrow$     & \multicolumn{1}{c}{\Large{0.475}} &  \multicolumn{1}{c}{\Large{0.790}}    & \multicolumn{1}{c|}{\Large{0.714}}   &  \multicolumn{1}{c}{\Large{0.801}}   &   \multicolumn{1}{c}{\Large{0.838}}    & \multicolumn{1}{c}{\Large{0.890}}  &\multicolumn{1}{c}{\Large{0.860}}  &  \multicolumn{1}{c}{\Large{0.886}}      &  \multicolumn{1}{c}{\Large{0.832}}     &    \multicolumn{1}{c|}{\textBC{red}{\Large{0.909}}}   &  \multicolumn{1}{c}{\Large{0.898}}     &   \multicolumn{1}{c}{\textBC{red}{\Large{0.911}}}    \\
&$\mathcal{M}\downarrow$ & \multicolumn{1}{c}{\Large{0.500}} &  \multicolumn{1}{c}{\Large{0.168}}    & \multicolumn{1}{c|}{\Large{0.219}}   &  \multicolumn{1}{c}{\Large{0.151}}   &   \multicolumn{1}{c}{\Large{0.100}}    & \multicolumn{1}{c}{\Large{0.063}}  &\multicolumn{1}{c}{\Large{0.082}}  &  \multicolumn{1}{c}{\Large{0.063}}      &  \multicolumn{1}{c}{\Large{0.082}}     &    \multicolumn{1}{c|}{\textBC{red}{\Large{0.050}}}   &  \multicolumn{1}{c}{\Large{0.059}}     &   \multicolumn{1}{c}{\textBC{red}{\Large{0.050}}}   \\

\hline
\multirow{6}{*}{\emph{\rotatebox{90}{NJUD~\cite{NJU2000}}}}      
&$F_{\beta}^{max}\uparrow$  & \multicolumn{1}{c}{\Large{0.328}} &  \multicolumn{1}{c}{\Large{0.769}}    & \multicolumn{1}{c|}{\Large{0.661}}   &  \multicolumn{1}{c}{\Large{0.789}}   &   \multicolumn{1}{c}{\Large{0.857}}    & \multicolumn{1}{c}{\Large{0.888}}  &\multicolumn{1}{c}{\Large{0.868}}  &  \multicolumn{1}{c}{\Large{0.888}}      &  \multicolumn{1}{c}{\Large{0.890}}     &    \multicolumn{1}{c|}{\textBC{red}{\Large{0.905}}}   &  \multicolumn{1}{c}{\Large{0.896}}     &   \multicolumn{1}{c}{\textBC{red}{\Large{0.910}}}    \\
&$F_{\beta}^{mean}\uparrow$  & \multicolumn{1}{c}{\Large{0.165}} &  \multicolumn{1}{c}{\Large{0.715}}    & \multicolumn{1}{c|}{\Large{0.618}}   &  \multicolumn{1}{c}{\Large{0.744}}   &   \multicolumn{1}{c}{\Large{0.788}}    & \multicolumn{1}{c}{\Large{0.844}}  &\multicolumn{1}{c}{\Large{0.813}}  &  \multicolumn{1}{c}{\Large{0.844}}      &  \multicolumn{1}{c}{\Large{0.837}}     &    \multicolumn{1}{c|}{\textBC{red}{\Large{0.877}}}   &  \multicolumn{1}{c}{\textBC{red}{\Large{0.871}}}     &   \multicolumn{1}{c}{\textBC{red}{\Large{0.871}}}      \\
&$F_{\beta}^{w}\uparrow$   & \multicolumn{1}{c}{\Large{0.234}} &  \multicolumn{1}{c}{\Large{0.497}}    & \multicolumn{1}{c|}{\Large{0.510}}   &  \multicolumn{1}{c}{\Large{0.545}}   &   \multicolumn{1}{c}{\Large{0.720}}    & \multicolumn{1}{c}{\Large{0.803}}  &\multicolumn{1}{c}{\Large{0.739}}  &  \multicolumn{1}{c}{\Large{0.805}}      &  \multicolumn{1}{c}{\Large{0.828}}     &    \multicolumn{1}{c|}{\textBC{red}{\Large{0.853}}}   &  \multicolumn{1}{c}{\Large{0.847}}     &   \multicolumn{1}{c}{\textBC{red}{\Large{0.857}}}       \\
& $S_m\uparrow$        & \multicolumn{1}{c}{\Large{0.413}} &  \multicolumn{1}{c}{\Large{0.703}}    & \multicolumn{1}{c|}{\Large{0.672}}   &  \multicolumn{1}{c}{\Large{0.735}}   &   \multicolumn{1}{c}{\Large{0.849}}    & \multicolumn{1}{c}{\Large{0.877}} &\multicolumn{1}{c}{\Large{0.859}}  &  \multicolumn{1}{c}{\Large{0.878}}      &  \multicolumn{1}{c}{\Large{0.878}}     &    \multicolumn{1}{c|}{\textBC{red}{\Large{0.897}}}   &  \multicolumn{1}{c}{\Large{0.885}}     &   \multicolumn{1}{c}{\textBC{red}{\Large{0.899}}}       \\
& $E_m\uparrow$     & \multicolumn{1}{c}{\Large{0.491}} &  \multicolumn{1}{c}{\Large{0.796}}    & \multicolumn{1}{c|}{\Large{0.751}}   &  \multicolumn{1}{c}{\Large{0.818}}   &   \multicolumn{1}{c}{\Large{0.866}}    & \multicolumn{1}{c}{\Large{0.909}}  &\multicolumn{1}{c}{\Large{0.882}}  &  \multicolumn{1}{c}{\Large{0.909}}      &  \multicolumn{1}{c}{\Large{0.900}}     &    \multicolumn{1}{c|}{\textBC{red}{\Large{0.926}}}   &  \multicolumn{1}{c}{\Large{0.920}}     &   \multicolumn{1}{c}{\textBC{red}{\Large{0.922}}}      \\
&$\mathcal{M}\downarrow$ & \multicolumn{1}{c}{\Large{0.448}} &  \multicolumn{1}{c}{\Large{0.167}}    & \multicolumn{1}{c|}{\Large{0.182}}   &  \multicolumn{1}{c}{\Large{0.151}}   &   \multicolumn{1}{c}{\Large{0.085}}    & \multicolumn{1}{c}{\Large{0.059}}  &\multicolumn{1}{c}{\Large{0.079}}  &  \multicolumn{1}{c}{\Large{0.061}}      &  \multicolumn{1}{c}{\Large{0.053}}     &    \multicolumn{1}{c|}{\textBC{red}{\Large{0.046}}}   &  \multicolumn{1}{c}{\Large{0.051}}     &   \multicolumn{1}{c}{\textBC{red}{\Large{0.045}}}      \\
\hline
\multirow{6}{*}{\emph{\rotatebox{90}{RGBD135~\cite{RGBD135}}}}      
&$F_{\beta}^{max}\uparrow$   & \multicolumn{1}{c}{\Large{0.800}} &  \multicolumn{1}{c}{\Large{0.311}}    & \multicolumn{1}{c|}{\Large{0.651}}   &  \multicolumn{1}{c}{\Large{0.625}}   &   \multicolumn{1}{c}{\Large{0.865}}    & \multicolumn{1}{c}{\Large{0.842}}  &\multicolumn{1}{c}{\Large{0.839}}  &  \multicolumn{1}{c}{\Large{0.853}}      &  \multicolumn{1}{c}{\Large{0.882}}     &    \multicolumn{1}{c|}{\textBC{red}{\Large{0.916}}}   &  \multicolumn{1}{c}{\Large{0.906}}     &   \multicolumn{1}{c}{\textBC{red}{\Large{0.928}}}      \\
&$F_{\beta}^{mean}\uparrow$  & \multicolumn{1}{c}{\Large{0.695}} &  \multicolumn{1}{c}{\Large{0.234}}    & \multicolumn{1}{c|}{\Large{0.594}}   &  \multicolumn{1}{c}{\Large{0.573}}   &   \multicolumn{1}{c}{\Large{0.778}}    & \multicolumn{1}{c}{\Large{0.774}}  &\multicolumn{1}{c}{\Large{0.762}}  &  \multicolumn{1}{c}{\Large{0.795}}      &  \multicolumn{1}{c}{\Large{0.829}}     &    \multicolumn{1}{c|}{\textBC{red}{\Large{0.891}}}   &  \multicolumn{1}{c}{\Large{0.867}}     &   \multicolumn{1}{c}{\textBC{red}{\Large{0.899}}}      \\
&$F_{\beta}^{w}\uparrow$   & \multicolumn{1}{c}{\Large{0.301}} &  \multicolumn{1}{c}{\Large{0.169}}    & \multicolumn{1}{c|}{\Large{0.478}}   &  \multicolumn{1}{c}{\Large{0.392}}   &   \multicolumn{1}{c}{\Large{0.687}}    & \multicolumn{1}{c}{\Large{0.711}}  &\multicolumn{1}{c}{\Large{0.650}}  &  \multicolumn{1}{c}{\Large{0.740}}      &  \multicolumn{1}{c}{\Large{0.787}}     &    \multicolumn{1}{c|}{\textBC{red}{\Large{0.848}}}   &  \multicolumn{1}{c}{\Large{0.843}}     &   \multicolumn{1}{c}{\textBC{red}{\Large{0.877}}}       \\
& $S_m\uparrow$        & \multicolumn{1}{c}{\Large{0.632}} &  \multicolumn{1}{c}{\Large{0.469}}    & \multicolumn{1}{c|}{\Large{0.709}}   &  \multicolumn{1}{c}{\Large{0.685}}   &   \multicolumn{1}{c}{\Large{0.863}}    & \multicolumn{1}{c}{\Large{0.843}} &\multicolumn{1}{c}{\Large{0.848}}  &  \multicolumn{1}{c}{\Large{0.858}}      &  \multicolumn{1}{c}{\Large{0.872}}     &    \multicolumn{1}{c|}{\textBC{red}{\Large{0.905}}}   &  \multicolumn{1}{c}{\Large{0.899}}     &   \multicolumn{1}{c}{\textBC{red}{\Large{0.924}}}        \\
& $E_m\uparrow$     & \multicolumn{1}{c}{\Large{0.817}} &  \multicolumn{1}{c}{\Large{0.676}}    & \multicolumn{1}{c|}{\Large{0.810}}   &  \multicolumn{1}{c}{\Large{0.806}}   &   \multicolumn{1}{c}{\Large{0.911}}    & \multicolumn{1}{c}{\Large{0.912}}  &\multicolumn{1}{c}{\Large{0.904}}  &  \multicolumn{1}{c}{\Large{0.919}}      &  \multicolumn{1}{c}{\Large{0.927}}     &    \multicolumn{1}{c|}{\textBC{red}{\Large{0.961}}}   &  \multicolumn{1}{c}{\Large{0.944}}     &   \multicolumn{1}{c}{\textBC{red}{\Large{0.968}}}       \\
&$\mathcal{M}\downarrow$ & \multicolumn{1}{c}{\Large{0.289}} &  \multicolumn{1}{c}{\Large{0.196}}    & \multicolumn{1}{c|}{\Large{0.120}}   &  \multicolumn{1}{c}{\Large{0.131}}   &   \multicolumn{1}{c}{\Large{0.055}}    & \multicolumn{1}{c}{\Large{0.050}}  &\multicolumn{1}{c}{\Large{0.065}}  &  \multicolumn{1}{c}{\Large{0.046}}      &  \multicolumn{1}{c}{\Large{0.038}}     &    \multicolumn{1}{c|}{\textBC{red}{\Large{0.028}}}   &  \multicolumn{1}{c}{\Large{0.030}}     &   \multicolumn{1}{c}{\textBC{red}{\Large{0.023}}}       \\

\hline
\multirow{6}{*}{\emph{\rotatebox{90}{DUTLF-D~\cite{DMRA}}}}      
&$F_{\beta}^{max}\uparrow$  & \multicolumn{1}{c}{\Large{0.770}} &  \multicolumn{1}{c}{\Large{0.444}}    & \multicolumn{1}{c|}{\Large{0.658}}   &  \multicolumn{1}{c}{\Large{0.774}}   &   \multicolumn{1}{c}{\Large{0.842}}    & \multicolumn{1}{c}{\Large{0.809}}  &\multicolumn{1}{c}{\Large{0.804}}  &  \multicolumn{1}{c}{\Large{0.823}}      &  \multicolumn{1}{c}{\Large{0.787}}     &    \multicolumn{1}{c|}{\textBC{red}{\Large{0.911}}}   &  \multicolumn{1}{c}{\Large{0.908}}     &   \multicolumn{1}{c}{\textBC{red}{\Large{0.918}}}  \\
&$F_{\beta}^{mean}\uparrow$ & \multicolumn{1}{c}{\Large{0.667}} &  \multicolumn{1}{c}{\Large{0.405}}    & \multicolumn{1}{c|}{\Large{0.633}}   &  \multicolumn{1}{c}{\Large{0.747}}   &   \multicolumn{1}{c}{\Large{0.792}}    & \multicolumn{1}{c}{\Large{0.760}}  &\multicolumn{1}{c}{\Large{0.753}}  &  \multicolumn{1}{c}{\Large{0.778}}      &  \multicolumn{1}{c}{\Large{0.735}}     &    \multicolumn{1}{c|}{\textBC{red}{\Large{0.884}}}   &  \multicolumn{1}{c}{\Large{0.883}}     &   \multicolumn{1}{c}{\textBC{red}{\Large{0.889}}}      \\
&$F_{\beta}^{w}\uparrow$   & \multicolumn{1}{c}{\Large{0.380}} &  \multicolumn{1}{c}{\Large{0.284}}    & \multicolumn{1}{c|}{\Large{0.521}}   &  \multicolumn{1}{c}{\Large{0.536}}   &   \multicolumn{1}{c}{\Large{0.682}}    & \multicolumn{1}{c}{\Large{0.688}}  &\multicolumn{1}{c}{\Large{0.628}}  &  \multicolumn{1}{c}{\Large{0.705}}      &  \multicolumn{1}{c}{\Large{0.638}}     &    \multicolumn{1}{c|}{\textBC{red}{\Large{0.847}}}   &  \multicolumn{1}{c}{\Large{0.852}}     &   \multicolumn{1}{c}{\textBC{red}{\Large{0.860}}}    \\
& $S_m\uparrow$        & \multicolumn{1}{c}{\Large{0.659}} &  \multicolumn{1}{c}{\Large{0.499}}    & \multicolumn{1}{c|}{\Large{0.687}}   &  \multicolumn{1}{c}{\Large{0.729}}   &   \multicolumn{1}{c}{\Large{0.831}}    & \multicolumn{1}{c}{\Large{0.801}} &\multicolumn{1}{c}{\Large{0.791}}  &  \multicolumn{1}{c}{\Large{0.808}}      &  \multicolumn{1}{c}{\Large{0.749}}     &    \multicolumn{1}{c|}{\textBC{red}{\Large{0.889}}}   &  \multicolumn{1}{c}{\Large{0.887}}     &   \multicolumn{1}{c}{\textBC{red}{\Large{0.899}}}     \\
& $E_m\uparrow$     & \multicolumn{1}{c}{\Large{0.751}} &  \multicolumn{1}{c}{\Large{0.712}}    & \multicolumn{1}{c|}{\Large{0.794}}   &  \multicolumn{1}{c}{\Large{0.842}}   &   \multicolumn{1}{c}{\Large{0.883}}    & \multicolumn{1}{c}{\Large{0.863}}  &\multicolumn{1}{c}{\Large{0.856}}  &  \multicolumn{1}{c}{\Large{0.871}}      &  \multicolumn{1}{c}{\Large{0.815}}     &    \multicolumn{1}{c|}{\textBC{red}{\Large{0.929}}}   &  \multicolumn{1}{c}{\Large{0.930}}     &   \multicolumn{1}{c}{\textBC{red}{\Large{0.937}}}       \\
&$\mathcal{M}\downarrow$ & \multicolumn{1}{c}{\Large{0.280}} &  \multicolumn{1}{c}{\Large{0.243}}    & \multicolumn{1}{c|}{\Large{0.159}}   &  \multicolumn{1}{c}{\Large{0.145}}   &   \multicolumn{1}{c}{\Large{0.097}}    & \multicolumn{1}{c}{\Large{0.100}}  &\multicolumn{1}{c}{\Large{0.112}}  &  \multicolumn{1}{c}{\Large{0.093}}      &  \multicolumn{1}{c}{\Large{0.100}}     &    \multicolumn{1}{c|}{\textBC{red}{\Large{0.047}}}   &  \multicolumn{1}{c}{\Large{0.048}}     &   \multicolumn{1}{c}{\textBC{red}{\Large{0.043}}}      \\

\hline
\multirow{6}{*}{\emph{\rotatebox{90}{NLPR~\cite{early_fusion_1}}}}      
&$F_{\beta}^{max}\uparrow$   & \multicolumn{1}{c}{\Large{0.695}} &  \multicolumn{1}{c}{\Large{0.413}}    & \multicolumn{1}{c|}{\Large{0.687}}   &  \multicolumn{1}{c}{\Large{0.752}}   &   \multicolumn{1}{c}{\Large{0.841}}    & \multicolumn{1}{c}{\Large{0.864}}  &\multicolumn{1}{c}{\Large{0.841}}  &  \multicolumn{1}{c}{\Large{0.876}}      &  \multicolumn{1}{c}{\Large{0.884}}     &    \multicolumn{1}{c|}{\textBC{red}{\Large{0.908}}} &  \multicolumn{1}{c}{\Large{0.888}}     &   \multicolumn{1}{c}{\textBC{red}{\Large{0.916}}}      \\
&$F_{\beta}^{mean}\uparrow$   & \multicolumn{1}{c}{\Large{0.583}} &  \multicolumn{1}{c}{\Large{0.328}}    & \multicolumn{1}{c|}{\Large{0.592}}   &  \multicolumn{1}{c}{\Large{0.683}}   &   \multicolumn{1}{c}{\Large{0.724}}    & \multicolumn{1}{c}{\Large{0.795}}  &\multicolumn{1}{c}{\Large{0.730}}  &  \multicolumn{1}{c}{\Large{0.796}}      &  \multicolumn{1}{c}{\Large{0.818}}     &    \multicolumn{1}{c|}{\textBC{red}{\Large{0.865}}}   &  \multicolumn{1}{c}{\Large{0.855}}     &   \multicolumn{1}{c}{\textBC{red}{\Large{0.870}}}     \\
&$F_{\beta}^{w}\uparrow$   & \multicolumn{1}{c}{\Large{0.254}} &  \multicolumn{1}{c}{\Large{0.259}}    & \multicolumn{1}{c|}{\Large{0.501}}   &  \multicolumn{1}{c}{\Large{0.516}}   &   \multicolumn{1}{c}{\Large{0.679}}    & \multicolumn{1}{c}{\Large{0.762}}  &\multicolumn{1}{c}{\Large{0.676}}  &  \multicolumn{1}{c}{\Large{0.780}}      &  \multicolumn{1}{c}{\Large{0.807}}     &    \multicolumn{1}{c|}{\textBC{red}{\Large{0.850}}}   &  \multicolumn{1}{c}{\Large{0.840}}     &   \multicolumn{1}{c}{\textBC{red}{\Large{0.862}}}       \\
& $S_m\uparrow  $        & \multicolumn{1}{c}{\Large{0.582}} &  \multicolumn{1}{c}{\Large{0.550}}    & \multicolumn{1}{c|}{\Large{0.724}}   &  \multicolumn{1}{c}{\Large{0.769}}   &   \multicolumn{1}{c}{\Large{0.860}}    & \multicolumn{1}{c}{\Large{0.874}}  &\multicolumn{1}{c}{\Large{0.856}}  &  \multicolumn{1}{c}{\Large{0.886}}      &  \multicolumn{1}{c}{\Large{0.884}}     &    \multicolumn{1}{c|}{\textBC{red}{\Large{0.908}}}   &  \multicolumn{1}{c}{\Large{0.898}}     &   \multicolumn{1}{c}{\textBC{red}{\Large{0.915}}}      \\
& $E_m\uparrow$         & \multicolumn{1}{c}{\Large{0.760}} &  \multicolumn{1}{c}{\Large{0.685}}    & \multicolumn{1}{c|}{\Large{0.786}}   &  \multicolumn{1}{c}{\Large{0.840}}   &   \multicolumn{1}{c}{\Large{0.869}}    & \multicolumn{1}{c}{\Large{0.916}}  &\multicolumn{1}{c}{\Large{0.872}}  &  \multicolumn{1}{c}{\Large{0.916}}      &  \multicolumn{1}{c}{\Large{0.920}}     &    \multicolumn{1}{c|}{\textBC{red}{\Large{0.945}}}   &  \multicolumn{1}{c}{\Large{0.942}}     &   \multicolumn{1}{c}{\textBC{red}{\Large{0.949}}}       \\
&$\mathcal{M}\downarrow$  & \multicolumn{1}{c}{\Large{0.301}} &  \multicolumn{1}{c}{\Large{0.196}}    & \multicolumn{1}{c|}{\Large{0.115}}   &  \multicolumn{1}{c}{\Large{0.100}}   &   \multicolumn{1}{c}{\Large{0.056}}    & \multicolumn{1}{c}{\Large{0.044}}  &\multicolumn{1}{c}{\Large{0.059}}  &  \multicolumn{1}{c}{\Large{0.041}}      &  \multicolumn{1}{c}{\Large{0.038}}     &    \multicolumn{1}{c|}{\textBC{red}{\Large{0.031}}}   &  \multicolumn{1}{c}{\Large{0.031}}     &   \multicolumn{1}{c}{\textBC{red}{\Large{0.028}}}      \\
\hline
\multirow{6}{*}{\emph{\rotatebox{90}{SIP~\cite{SIP}}}}      
&$F_{\beta}^{max}\uparrow$  & \multicolumn{1}{c}{\Large{0.720}} &  \multicolumn{1}{c}{\Large{0.680}}    & \multicolumn{1}{c|}{\Large{0.544}}   &  \multicolumn{1}{c}{\Large{0.704}}   &   \multicolumn{1}{c}{\Large{0.720}}    & \multicolumn{1}{c}{\Large{0.861}}  &\multicolumn{1}{c}{\Large{0.840}}  &  \multicolumn{1}{c}{\Large{0.851}}      &  \multicolumn{1}{c}{\Large{0.870}}     &    \multicolumn{1}{c|}{\textBC{red}{\Large{0.901}}}   &  \multicolumn{1}{c}{\Large{0.847}}     &   \multicolumn{1}{c}{\textBC{red}{\Large{0.892}}}     \\
&$F_{\beta}^{mean}\uparrow$  & \multicolumn{1}{c}{\Large{0.644}} &  \multicolumn{1}{c}{\Large{0.645}}    & \multicolumn{1}{c|}{\Large{0.495}}   &  \multicolumn{1}{c}{\Large{0.673}}   &   \multicolumn{1}{c}{\Large{0.684}}    & \multicolumn{1}{c}{\Large{0.825}}  &\multicolumn{1}{c}{\Large{0.795}}  &  \multicolumn{1}{c}{\Large{0.809}}      &  \multicolumn{1}{c}{\Large{0.819}}     &    \multicolumn{1}{c|}{\textBC{red}{\Large{0.864}}}   &  \multicolumn{1}{c}{\Large{0.815}}     &   \multicolumn{1}{c}{\textBC{red}{\Large{0.855}}}      \\
&$F_{\beta}^{w}\uparrow$   & \multicolumn{1}{c}{\Large{0.342}} &  \multicolumn{1}{c}{\Large{0.414}}    & \multicolumn{1}{c|}{\Large{0.397}}   &  \multicolumn{1}{c}{\Large{0.406}}   &   \multicolumn{1}{c}{\Large{0.535}}    & \multicolumn{1}{c}{\Large{0.768}}  &\multicolumn{1}{c}{\Large{0.712}}  &  \multicolumn{1}{c}{\Large{0.748}}      &  \multicolumn{1}{c}{\Large{0.788}}     &    \multicolumn{1}{c|}{\textBC{red}{\Large{0.829}}}   &  \multicolumn{1}{c}{\Large{0.734}}     &   \multicolumn{1}{c}{\textBC{red}{\Large{0.822}}}      \\
& $S_m\uparrow$        & \multicolumn{1}{c}{\Large{0.616}} &  \multicolumn{1}{c}{\Large{0.683}}    & \multicolumn{1}{c|}{\Large{0.595}}   &  \multicolumn{1}{c}{\Large{0.653}}   &   \multicolumn{1}{c}{\Large{0.716}}    & \multicolumn{1}{c}{\Large{0.842}} &\multicolumn{1}{c}{\Large{0.833}}  &  \multicolumn{1}{c}{\Large{0.835}}      &  \multicolumn{1}{c}{\Large{0.850}}     &    \multicolumn{1}{c|}{\textBC{red}{\Large{0.878}}}   &  \multicolumn{1}{c}{\Large{0.800}}     &   \multicolumn{1}{c}{\textBC{red}{\Large{0.875}}}       \\
& $E_m\uparrow$     & \multicolumn{1}{c}{\Large{0.751}} &  \multicolumn{1}{c}{\Large{0.787}}    & \multicolumn{1}{c|}{\Large{0.722}}   &  \multicolumn{1}{c}{\Large{0.794}}   &   \multicolumn{1}{c}{\Large{0.824}}    & \multicolumn{1}{c}{\Large{0.900}}  &\multicolumn{1}{c}{\Large{0.886}}  &  \multicolumn{1}{c}{\Large{0.894}}      &  \multicolumn{1}{c}{\Large{0.899}}     &    \multicolumn{1}{c|}{\textBC{red}{\Large{0.914}}}   &  \multicolumn{1}{c}{\Large{0.858}}     &   \multicolumn{1}{c}{\textBC{red}{\Large{0.915}}}       \\
&$\mathcal{M}\downarrow$ & \multicolumn{1}{c}{\Large{0.298}} &  \multicolumn{1}{c}{\Large{0.186}}    & \multicolumn{1}{c|}{\Large{0.224}}   &  \multicolumn{1}{c}{\Large{0.185}}   &   \multicolumn{1}{c}{\Large{0.139}}    & \multicolumn{1}{c}{\Large{0.071}}  &\multicolumn{1}{c}{\Large{0.086}}  &  \multicolumn{1}{c}{\Large{0.075}}      &  \multicolumn{1}{c}{\Large{0.064}}     &    \multicolumn{1}{c|}{\textBC{red}{\Large{0.054}}}   &  \multicolumn{1}{c}{\Large{0.088}}     &   \multicolumn{1}{c}{\textBC{red}{\Large{0.054}}}      \\
   \bottomrule[2pt]
   \end{tabular}
 }
  \end{table*}
  
\begin{table*}[!ht]
\caption{The model sizes and average speed of different methods.}
\centering
\label{tab:model-size}
\resizebox{\textwidth}{!}  
{\begin{tabu}{cccccccc}
\toprule[2pt]
    Model Name & PCANet~\cite{PCA} & MMCI~\cite{MMCI} & TANet~\cite{TANet} & CPFP~\cite{CPFP} &DMRA~\cite{DMRA}  & OURS(VGG-19) & OURS(VGG-16)\\
\midrule[1pt]
Model Size& 533.6 (MB)      & 951.9 (MB)             & 929.7 (MB)             & 278 (MB)                & 238.8 (MB)                 &128.1 (MB)  &106.7 (MB)   \\
\midrule[1pt]
Average speed& 17 (FPS)      & 20 (FPS)             & 14(FPS)             & 6 (FPS)                & 22 (FPS)  &30 (FPS)  &32 (FPS)   \\

\bottomrule[2pt]
\end{tabu}
% \vspace{-5.5mm}
}
\end{table*}

\textbf{Quantitative Evaluation.}
 1) Tab.~\ref{tab:scores} shows performance comparisons in terms of the maximum F-measure, mean F-measure, weighted F-measure, S-measure, E-measure and MAE scores. 
It can be seen that our DANet achieves the best results on all six datasets  under all six metrics. 
2) Tab.~\ref{tab:model-size} lists the model sizes and average speed of different methods in detail. Our model is the smallest and the fastest among these state-of-art methods and saves $55.5\%$ of the parameters compared to the second lightest method DMRA~\cite{DMRA} .
\begin{figure*}
  \centering
  \includegraphics[width=\textwidth]{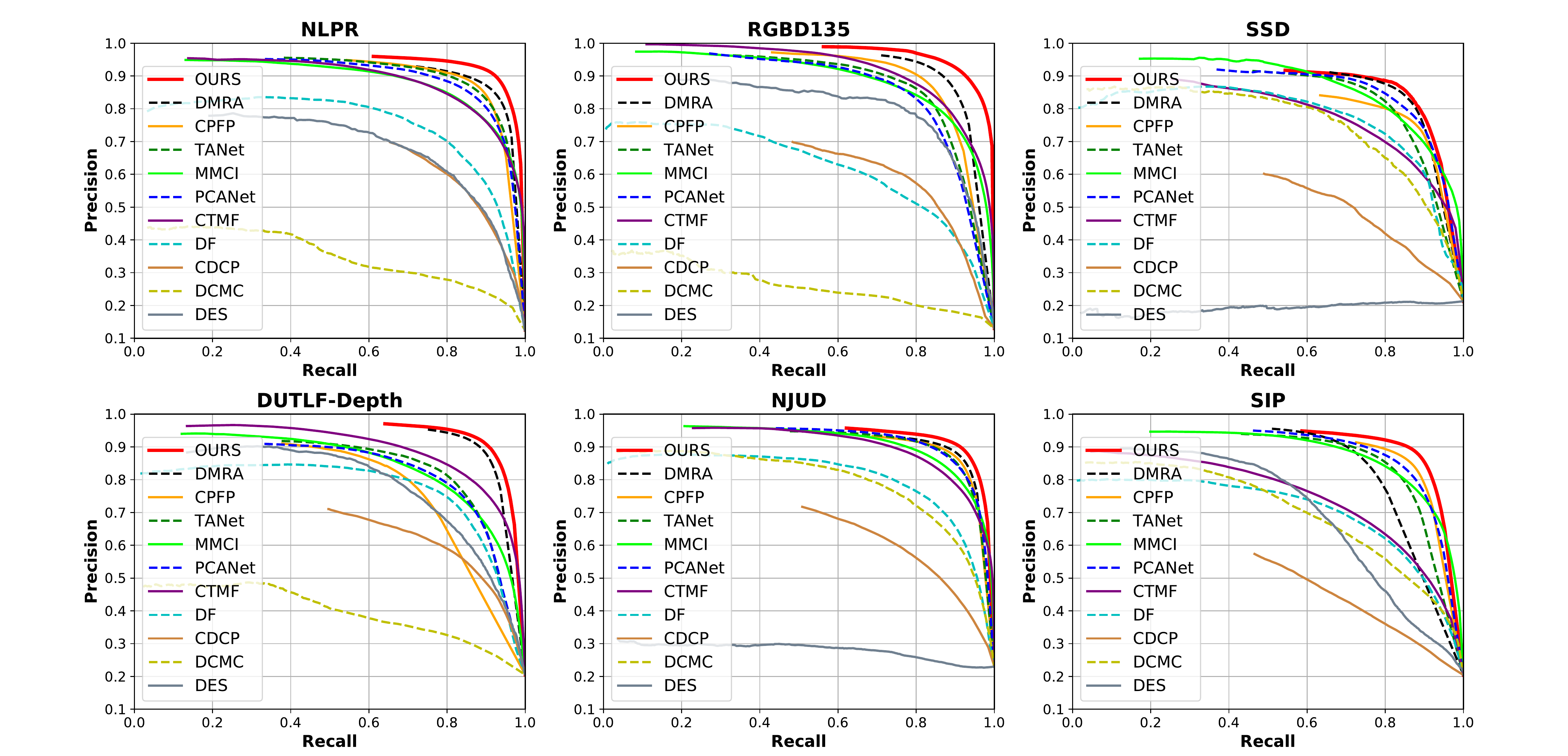}
  \caption{Precision (vertical axis) recall (horizontal axis) curves on six RGB-D salient object detection datasets.}
  \label{fig:PR}
  \end{figure*} 
3) Fig.~\ref{fig:PR} shows the PR curves of different algorithms. We can see that the curves of the proposed method are significantly higher than those of other methods, especially on the NJUD, NLPR and RGBD135 datasets which contain plenty of relatively complex images.  
Through detailed quantitative comparisons, it can be seen that our method has significant advantages in accuracy and model size, which indicates it is necessary to further explore how to better utilize depth information. 

\begin{figure*}[ht]
  \centering
  \includegraphics[width=\textwidth,height=0.65\linewidth]{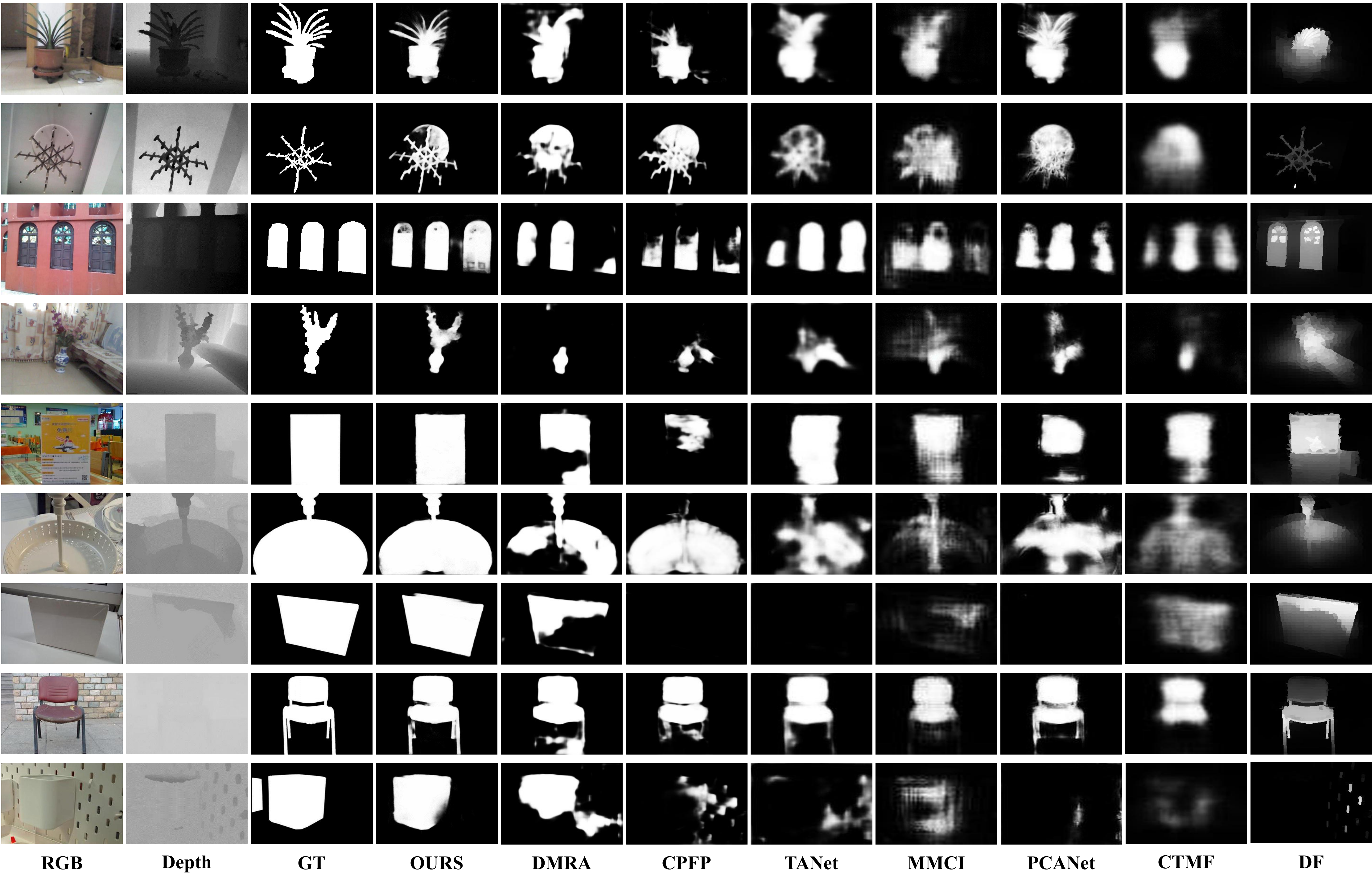}
  \caption{Visual comparison between our results and the state-of-the-art methods.}
  \label{fig:visual-cmp}
\end{figure*}

 \textbf{Qualitative Evaluation.}
 Fig.~\ref{fig:visual-cmp} illustrates the visual comparison with other approaches. Our method yields the results more close to the ground truth in various challenging scenarios. %
For example, for the images having multiple objects or the objects having slender parts, our method can accurately locate objects and capture more details (see the $1^{st}$ - $3^{th}$ rows). In complex environments, with the guidance of the depth maps, the proposed method can precisely identify the whole object, while other methods fail (see the $4^{th}$ - $6^{th}$ rows). Even when the depth information performs badly in separating the foreground from the background, our network still significantly outperforms other methods (see the $7^{th}$ - $9^{th}$ rows).

\begin{table*}[!ht]
\scriptsize
  \caption{
  Ablation analysis on five datasets.
  }
  \label{tab:rgbd_performance}
   \renewcommand\tabcolsep{5.0pt} % 调整表格列间的宽度
   \renewcommand\arraystretch{1.5}
  \centering

\resizebox{0.72\textwidth}{!}  
   %\scalebox{.84}
  {
\begin{tabular}{ll|l|lll|llll|ll}

   \toprule[2pt]
 
 \multicolumn{2}{l|}{\makecell[c]{Metric}}  & \makecell[c]\small{\textbf{Baseline}}   & \makecell[c]\Large{Add$_p$}         &\makecell[c]\Large{Add$_{He}$}     &\makecell[c]\Large{\textbf{Cat$_{He}$}} &\makecell[c]\Large{DA} & \makecell[c]\Large{MGA} & \makecell[c]\Large{DEFA} & \makecell[c]\Large{\textbf{DEDA}} &\makecell[c]\Large{ASPP} &\makecell[c] \Large{\textbf{PAFE}}  \\
\hline
\multirow{5}{*}{\emph{\rotatebox{90}{SSD~\cite{SSD}}}}      
&$F_{\beta}^{max}\uparrow$   & \makecell[c]{\small{0.799}} &  \makecell[c]{\small{0.812}}    & \makecell[c]{\small{0.817}}   &  \makecell[c]{\textBC{red}{\small{0.845}}}   &   \makecell[c]{\small{0.837}}    &\makecell[c]{\small{0.843}}  &\makecell[c]{\small{0.858}} &\makecell[c]{\textBC{red}{\small{0.860}}}  & \makecell[c]{\small{0.879}}      &  \makecell[c]{\textBC{red}{\small{0.888}}}     \\
&$F_{\beta}^{mean}\uparrow$   & \makecell[c]{\small{0.745}} &  \makecell[c]{\small{0.743}}    & \makecell[c]{\small{0.734}}   &  \makecell[c]{\textBC{red}{\small{0.758}}}   &   \makecell[c]{\small{0.754}}    &\makecell[c]{\small{0.794}} &\makecell[c]{\small{0.806}} &\makecell[c]{\textBC{red}{\small{0.810}}}  & \makecell[c]{\small{0.830}}      &  \makecell[c]{\textBC{red}{\small{0.831}}}   \\
&$F_{\beta}^{w}\uparrow$   & \makecell[c]{\small{0.700}} &  \makecell[c]{\small{0.705}}    & \makecell[c]{\small{0.677}}   &  \makecell[c]{\textBC{red}{\small{0.710}}}   &   \makecell[c]{\small{0.697}}    &\makecell[c]{\small{0.745}}  &\makecell[c]{\small{0.757}} &\makecell[c]{\textBC{red}{\small{0.761}}}  &
\makecell[c]{\small{0.784}}      &  \makecell[c]{\textBC{red}{\small{0.798}}}   \\
& $S_m\uparrow  $        & \makecell[c]{\small{0.813}} &  \makecell[c]{\small{0.825}}    & \makecell[c]{\small{0.811}}   &  \makecell[c]{\textBC{red}{\small{0.835}}}   &   \makecell[c]{\small{0.829}}    &\makecell[c]{\small{0.841}}  &\makecell[c]{\small{0.846}} &\makecell[c]{\textBC{red}{\small{0.847}}}  & \makecell[c]{\small{0.855}}      &  \makecell[c]{\textBC{red}{\small{0.869}}}    \\
& $E_m\uparrow$         & \makecell[c]{\small{0.862}} &  \makecell[c]{\textBC{red}{\small{0.857}}}    & \makecell[c]{\small{0.833}}   &  \makecell[c]{\small{0.849}}   &   \makecell[c]{\small{0.847}}    &\makecell[c]{\small{0.883}} &\makecell[c]{\small{0.886}}  &\makecell[c]{\textBC{red}{\small{0.887}}}  & \makecell[c]{\small{0.905}}      &  \makecell[c]{\textBC{red}{\small{0.909}}}   \\
&$\mathcal{M}\downarrow$  & \makecell[c]{\small{0.080}} &  \makecell[c]{\small{0.077}}    & \makecell[c]{\small{0.092}}   &  \makecell[c]{\textBC{red}{\small{0.076}}}   &   \makecell[c]{\small{0.078}}    &\makecell[c]{\small{0.064}}  &\makecell[c]{\small{0.060}} &\makecell[c]{\textBC{red}{\small{0.062}}}  & \makecell[c]{\small{0.056}}      &  \makecell[c]{\textBC{red}{\small{0.050}}}   \\
\hline

\multirow{5}{*}{\emph{\rotatebox{90}{NJUD~\cite{NJU2000}}}}      
&$F_{\beta}^{max}\uparrow$   & \makecell[c]{\small{0.855}} &  \makecell[c]{\small{0.861}}    & \makecell[c]{\small{0.857}}   &  \makecell[c]{\textBC{red}{\small{0.869}}}   &   \makecell[c]{\small{0.865}}    &\makecell[c]{\small{0.882}}  &\makecell[c]{\small{0.889}} &\makecell[c]{\textBC{red}{\small{0.889}}}  & \makecell[c]{\small{0.896}}      &  \makecell[c]{\textBC{red}{\small{0.905}}}     \\
&$F_{\beta}^{mean}\uparrow$   & \makecell[c]{\small{0.781}} &  \makecell[c]{\small{0.784}}    & \makecell[c]{\small{0.798}}   &  \makecell[c]{\textBC{red}{\small{0.815}}}   &   \makecell[c]{\small{0.813}}    &\makecell[c]{\small{0.832}}  &\makecell[c]{\small{0.842}} &\makecell[c]{\textBC{red}{\small{0.849}}}  & \makecell[c]{\small{0.862}}      &  \makecell[c]{\textBC{red}{\small{0.877}}}   \\
&$F_{\beta}^{w}\uparrow$   & \makecell[c]{\small{0.748}} &  \makecell[c]{\small{0.757}}    & \makecell[c]{\small{0.744}}   &  \makecell[c]{\textBC{red}{\small{0.770}}}   &   \makecell[c]{\small{0.763}}    &\makecell[c]{\small{0.815}} &\makecell[c]{\small{0.823}}  &\makecell[c]{\textBC{red}{\small{0.826}}}  & \makecell[c]{\small{0.843}}      &  \makecell[c]{\textBC{red}{\small{0.853}}}   \\
& $S_m\uparrow  $        & \makecell[c]{\small{0.848}} &  \makecell[c]{\small{0.854}}    & \makecell[c]{\small{0.847}}   &  \makecell[c]{\textBC{red}{\small{0.860}}}   &   \makecell[c]{\small{0.856}}    &\makecell[c]{\small{0.878}}  &\makecell[c]{\textBC{red}{\small{0.881}}} &\makecell[c]{\small{0.880}}  & \makecell[c]{\small{0.890}}      &  \makecell[c]{\textBC{red}{\small{0.897}}}    \\
& $E_m\uparrow$         & \makecell[c]{\small{0.863}} &  \makecell[c]{\small{0.866}}    & \makecell[c]{\small{0.872}}   &  \makecell[c]{\textBC{red}{\small{0.880}}}   &   \makecell[c]{\small{0.880}}    &\makecell[c]{\small{0.896}}  &\makecell[c]{\small{0.904}} &\makecell[c]{\textBC{red}{\small{0.907}}}  & \makecell[c]{\small{0.915}}      &  \makecell[c]{\textBC{red}{\small{0.926}}}   \\
&$\mathcal{M}\downarrow$  & \makecell[c]{\small{0.079}} &  \makecell[c]{\small{0.076}}    & \makecell[c]{\small{0.081}}   &  \makecell[c]{\textBC{red}{\small{0.073}}}   &   \makecell[c]{\small{0.076}}    &\makecell[c]{\small{0.059}} &\makecell[c]{\small{0.056}}  &\makecell[c]{\textBC{red}{\small{0.055}}}  & \makecell[c]{\small{0.049}}      &  \makecell[c]{\textBC{red}{\small{0.046}}}\\

\hline

\multirow{5}{*}{\emph{\rotatebox{90}{RGBD135~\cite{RGBD135}}}}      
&$F_{\beta}^{max}\uparrow$   & \makecell[c]{\small{0.839}} &  \makecell[c]{\small{0.860}}    & \makecell[c]{\small{0.865}}   &  \makecell[c]{\textBC{red}{\small{0.877}}}   &   \makecell[c]{\small{0.881}}    &\makecell[c]{\small{0.897}} &\makecell[c]{\small{0.904}}  &\makecell[c]{\textBC{red}{\small{0.913}}}  & \makecell[c]{\small{0.907}}      &  \makecell[c]{\textBC{red}{\small{0.916}}}     \\
&$F_{\beta}^{mean}\uparrow$   & \makecell[c]{\small{0.772}} &  \makecell[c]{\small{0.792}}    & \makecell[c]{\small{0.802}}   &  \makecell[c]{\textBC{red}{\small{0.814}}}   &   \makecell[c]{\small{0.812}}    &\makecell[c]{\small{0.850}} &\makecell[c]{\small{0.868}}  &\makecell[c]{\textBC{red}{\small{0.876}}}  & \makecell[c]{\textBC{red}{\small{0.894}}}      &  \makecell[c]{\small{0.891}}   \\
&$F_{\beta}^{w}\uparrow$   & \makecell[c]{\small{0.705}} &  \makecell[c]{\small{0.732}}    & \makecell[c]{\small{0.740}}   &  \makecell[c]{\textBC{red}{\small{0.742}}}   &   \makecell[c]{\small{0.751}}    &\makecell[c]{\small{0.823}}  &\makecell[c]{\small{0.831}} &\makecell[c]{\textBC{red}{\small{0.846}}}  & \makecell[c]{\textBC{red}{\small{0.860}}}      &  \makecell[c]{\small{0.848}}   \\
& $S_m\uparrow  $        & \makecell[c]{\small{0.847}} &  \makecell[c]{\small{0.863}}    & \makecell[c]{\textBC{red}{\small{0.867}}}   &  \makecell[c]{\small{0.864}}   &   \makecell[c]{\small{0.871}}    &\makecell[c]{\small{0.906}}  &\makecell[c]{\small{0.903}} &\makecell[c]{\textBC{red}{\small{0.907}}}  & \makecell[c]{\textBC{red}{\small{0.915}}}      &  \makecell[c]{\small{0.905}}    \\
& $E_m\uparrow$         & \makecell[c]{\small{0.904}} &  \makecell[c]{\small{0.910}}    & \makecell[c]{\small{0.922}}   &  \makecell[c]{\textBC{red}{\small{0.922}}}   &   \makecell[c]{\small{0.923}}    &\makecell[c]{\small{0.943}}  &\makecell[c]{\small{0.952}} &\makecell[c]{\textBC{red}{\small{0.954}}}  & \makecell[c]{\textBC{red}{\small{0.966}}}      &  \makecell[c]{\small{0.961}}   \\
&$\mathcal{M}\downarrow$  & \makecell[c]{\small{0.055}} &  \makecell[c]{\small{0.050}}    & \makecell[c]{\small{0.051}}   &  \makecell[c]{\textBC{red}{\small{0.046}}}   &   \makecell[c]{\small{0.044}}    &\makecell[c]{\small{0.032}}  &\makecell[c]{\small{0.033}} &\makecell[c]{\textBC{red}{\small{0.029}}}  & \makecell[c]{\textBC{red}{\small{0.026}}}      &  \makecell[c]{\small{0.028}}\\
\hline

\multirow{5}{*}{\emph{\rotatebox{90}{NLPR~\cite{early_fusion_1}}}}      
&$F_{\beta}^{max}\uparrow$   & \makecell[c]{\small{0.852}} &  \makecell[c]{\small{0.852}}    & \makecell[c]{\small{0.860}}   &  \makecell[c]{\textBC{red}{\small{0.862}}}   &   \makecell[c]{\small{0.859}}    &\makecell[c]{\textBC{red}{\small{0.887}}}  &\makecell[c]{\small{0.886}} &\makecell[c]{\small{0.880}}  & \makecell[c]{\small{0.903}}      &  \makecell[c]{\small{\textBC{red}{\small{0.908}}}}     \\
&$F_{\beta}^{mean}\uparrow$   & \makecell[c]{\small{0.772}} &  \makecell[c]{\small{0.772}}    & \makecell[c]{\small{0.773}}   &  \makecell[c]{\textBC{red}{\small{0.774}}}   &   \makecell[c]{\small{0.773}}    &\makecell[c]{\small{0.821}}  &\makecell[c]{\small{0.826}} &\makecell[c]{\textBC{red}{\small{0.832}}}  & \makecell[c]{\small{0.857}}      &  \makecell[c]{\textBC{red}{\small{0.865}}}   \\
&$F_{\beta}^{w}\uparrow$   & \makecell[c]{\small{0.741}} &  \makecell[c]{\small{0.741}}    & \makecell[c]{\small{0.743}}   &  \makecell[c]{\textBC{red}{\small{0.743}}}   &   \makecell[c]{\small{0.734}}    &\makecell[c]{\small{0.809}}  &\makecell[c]{\small{0.813}} &\makecell[c]{\textBC{red}{\small{0.815}}}  & \makecell[c]{\small{0.846}}      &  \makecell[c]{\textBC{red}{\small{0.850}}}   \\
& $S_m\uparrow  $        & \makecell[c]{\small{0.862}} &  \makecell[c]{\small{0.863}}    & \makecell[c]{\small{0.866}}   &  \makecell[c]{\textBC{red}{\small{0.868}}}   &   \makecell[c]{\small{0.865}}    &\makecell[c]{\textBC{red}{\small{0.893}}} &\makecell[c]{\textBC{red}{\small{0.893}}}  &\makecell[c]{\small{0.889}}  & \makecell[c]{\small{0.907}}      &  \makecell[c]{\textBC{red}{\small{0.908}}}    \\
& $E_m\uparrow$         & \makecell[c]{\small{0.898}} &  \makecell[c]{\textBC{red}{\small{0.900}}}    & \makecell[c]{\small{0.898}}   &  \makecell[c]{\small{0.892}}   &   \makecell[c]{\small{0.894}}    &\makecell[c]{\small{0.920}}  &\makecell[c]{\small{0.923}} &\makecell[c]{\textBC{red}{\small{0.926}}}  & \makecell[c]{\small{0.939}}      &  \makecell[c]{\textBC{red}{\small{0.945}}}   \\
&$\mathcal{M}\downarrow$  & \makecell[c]{\small{0.052}} &  \makecell[c]{\small{0.053}}    & \makecell[c]{\small{0.053}}   &  \makecell[c]{\textBC{red}{\small{0.052}}}   &   \makecell[c]{\small{0.055}}    &\makecell[c]{\small{0.040}}  &\makecell[c]{\small{0.040}} &\makecell[c]{\textBC{red}{\small{0.038}}}  & \makecell[c]{\small{0.032}}      &  \makecell[c]{\textBC{red}{\small{0.031}}}\\
\hline

\multirow{5}{*}{\emph{\rotatebox{90}{SIP~\cite{SIP}}}}      
&$F_{\beta}^{max}\uparrow$   & \makecell[c]{\small{0.838}} &  \makecell[c]{\small{0.851}}    & \makecell[c]{\small{0.836}}   &  \makecell[c]{\textBC{red}{\small{0.849}}}   &   \makecell[c]{\small{0.835}}    &\makecell[c]{\small{0.864}}  &\makecell[c]{\small{0.873}} &\makecell[c]{\textBC{red}{\small{0.876}}}  & \makecell[c]{\small{0.885}}      &  \makecell[c]{\textBC{red}{\small{0.901}}}     \\
&$F_{\beta}^{mean}\uparrow$   & \makecell[c]{\small{0.780}} &  \makecell[c]{\small{0.784}}    & \makecell[c]{\small{0.758}}   &  \makecell[c]{\textBC{red}{\small{0.787}}}   &   \makecell[c]{\small{0.771}}    &\makecell[c]{\small{0.804}}  &\makecell[c]{\small{0.830}} &\makecell[c]{\textBC{red}{\small{0.833}}}  & \makecell[c]{\small{0.847}}      &  \makecell[c]{\textBC{red}{\small{0.864}}}   \\
&$F_{\beta}^{w}\uparrow$   & \makecell[c]{\small{0.716}} &  \makecell[c]{\small{0.721}}    & \makecell[c]{\small{0.692}}   &  \makecell[c]{\textBC{red}{\small{0.722}}}   &   \makecell[c]{\small{0.699}}    &\makecell[c]{\small{0.767}}  &\makecell[c]{\small{0.791}} &\makecell[c]{\textBC{red}{\small{0.798}}}  & \makecell[c]{\small{0.813}}      &  \makecell[c]{\textBC{red}{\small{0.829}}}   \\
& $S_m\uparrow  $        & \makecell[c]{\small{0.833}} &  \makecell[c]{\small{0.840}}    & \makecell[c]{\small{0.824}}   &  \makecell[c]{\textBC{red}{\small{0.841}}}   &   \makecell[c]{\small{0.833}}    &\makecell[c]{\small{0.854}}  &\makecell[c]{\small{0.863}} &\makecell[c]{\textBC{red}{\small{0.865}}}  & \makecell[c]{\small{0.871}}      &  \makecell[c]{\textBC{red}{\small{0.878}}}    \\
& $E_m\uparrow$         & \makecell[c]{\small{0.882}} &  \makecell[c]{\textBC{red}{\small{0.881}}}    & \makecell[c]{\small{0.867}}   &  \makecell[c]{\small{0.880}}   &   \makecell[c]{\small{0.868}}    &\makecell[c]{\small{0.889}}  &\makecell[c]{\textBC{red}{\small{0.907}}} &\makecell[c]{\textBC{red}{\small{0.907}}}  & \makecell[c]{\small{0.909}}      &  \makecell[c]{\textBC{red}{\small{0.917}}}   \\
&$\mathcal{M}\downarrow$  & \makecell[c]{\small{0.085}} &  \makecell[c]{\textBC{red}{\small{0.082}}}    & \makecell[c]{\small{0.095}}   &  \makecell[c]{\small{0.083}}   &   \makecell[c]{\small{0.092}}    &\makecell[c]{\small{0.070}}  &\makecell[c]{\small{0.062}}  &\makecell[c]{\textBC{red}{\small{0.061}}}  & \makecell[c]{\small{0.057}}      &  \makecell[c]{\textBC{red}{\small{0.054}}} \\
   \bottomrule[2pt]
   \label{tab:ablation}
   \end{tabular}
 }
  \end{table*}
  
 \begin{figure}
  \centering
  \includegraphics[width=0.8\linewidth,height=0.4\linewidth]{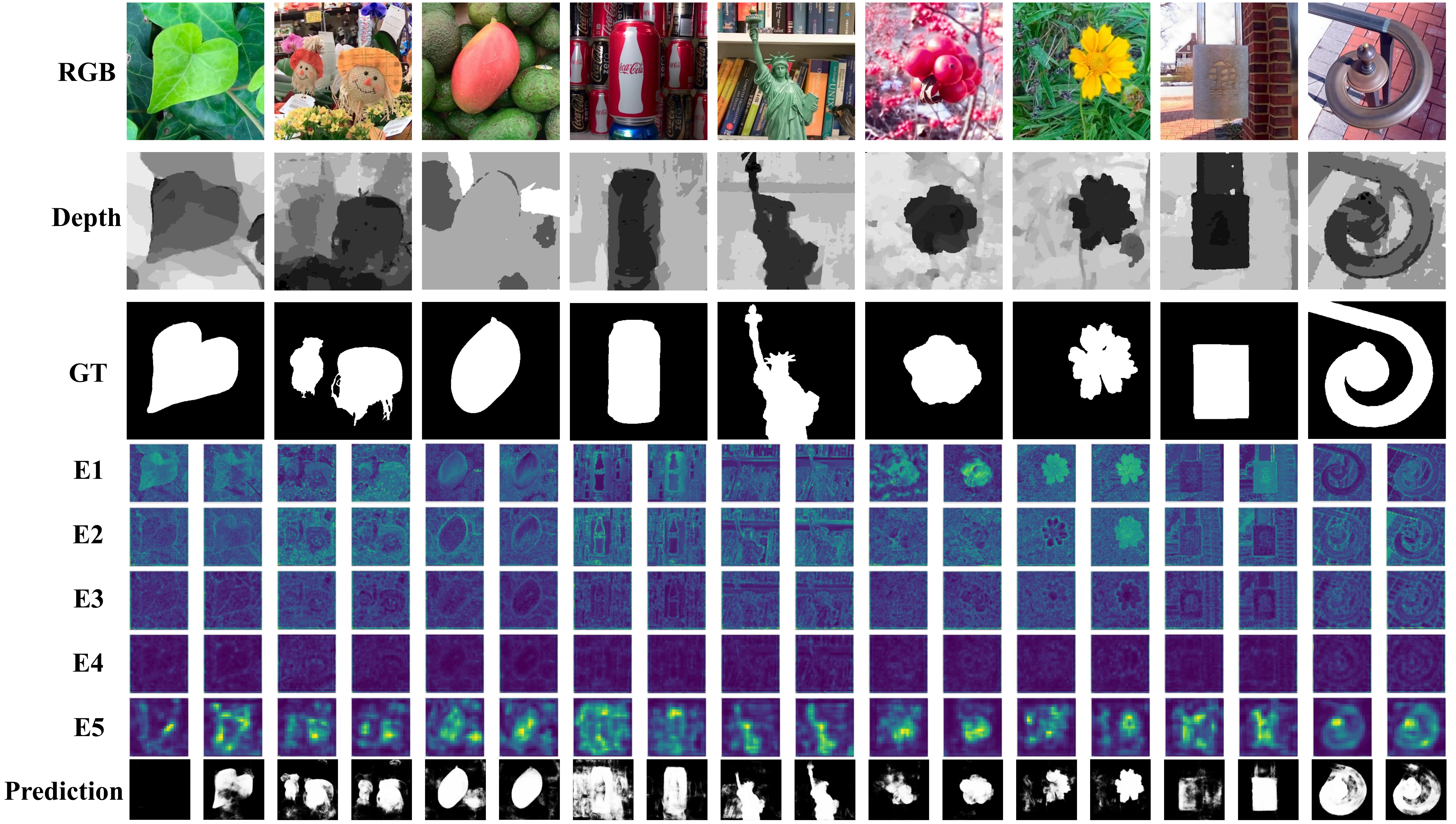}
  \caption{Visual comparison between the 4-channel RGB-D FPN and the 3-channel RGB FPN (baseline). Each input image corresponds to two columns of feature maps ($\mathbf{E}^1 \sim \mathbf{E}^5$) and prediction. The left is the results of the 3-channel baseline, while the right is those of the 4-channel baseline. 
  }\label{fig:rgb_rgbd_fpn}
  \end{figure}

\subsection{Ablation Studies}
We take the FPN network of the VGG-16 backbone as the baseline to analyze the contribution of each component. To verify their generalization abilities, we demonstrate the experimental results on five datasets.

\textbf{Effectiveness of Depth Fusion in Encoder Network.} We evaluate three early fusion strategies. The results are shown in Tab.~\ref{tab:ablation}. Add$_p$ denotes the fusion by using element-wise addition and the ImageNet pre-trained first-layer convolution. Add$_{He}$ and Cat$_{He}$ use the He's initialization~\cite{PRelu} instead of the pre-trained parameters in the first layer, and the latter adopts the 4-channel concatenation rather than element-wise addition. 
We can see that Cat$_{He}$ is significantly better than the baseline and other early fusion methods across five datasets. In particular, it respectively achieves the gain of $4.53\%$, $5.44\%$, $5.25\%$ and $16.36\%$ in terms of the $F_{\beta}^{max}$, $F_{\beta}^{mean}$, $F_{\beta}^{w}$ and MAE on the RGBD135 dataset.
Furthermore, we visualize the features of different levels in Fig.~\ref{fig:rgb_rgbd_fpn}. With the aid of the contrast prior provided the depth map, salient objects and their surrounding backgrounds can be clearly distinguished starting from the lowest level ($\mathbf{E}^1$). At the highest level ($\mathbf{E}^5$), the encoder feature is more concentrated on the salient regions, thereby providing the decoder with effective contextual guidance. 

\begin{figure}[!ht]
  \centering
  \includegraphics[width=0.6\linewidth,height=0.4\linewidth]{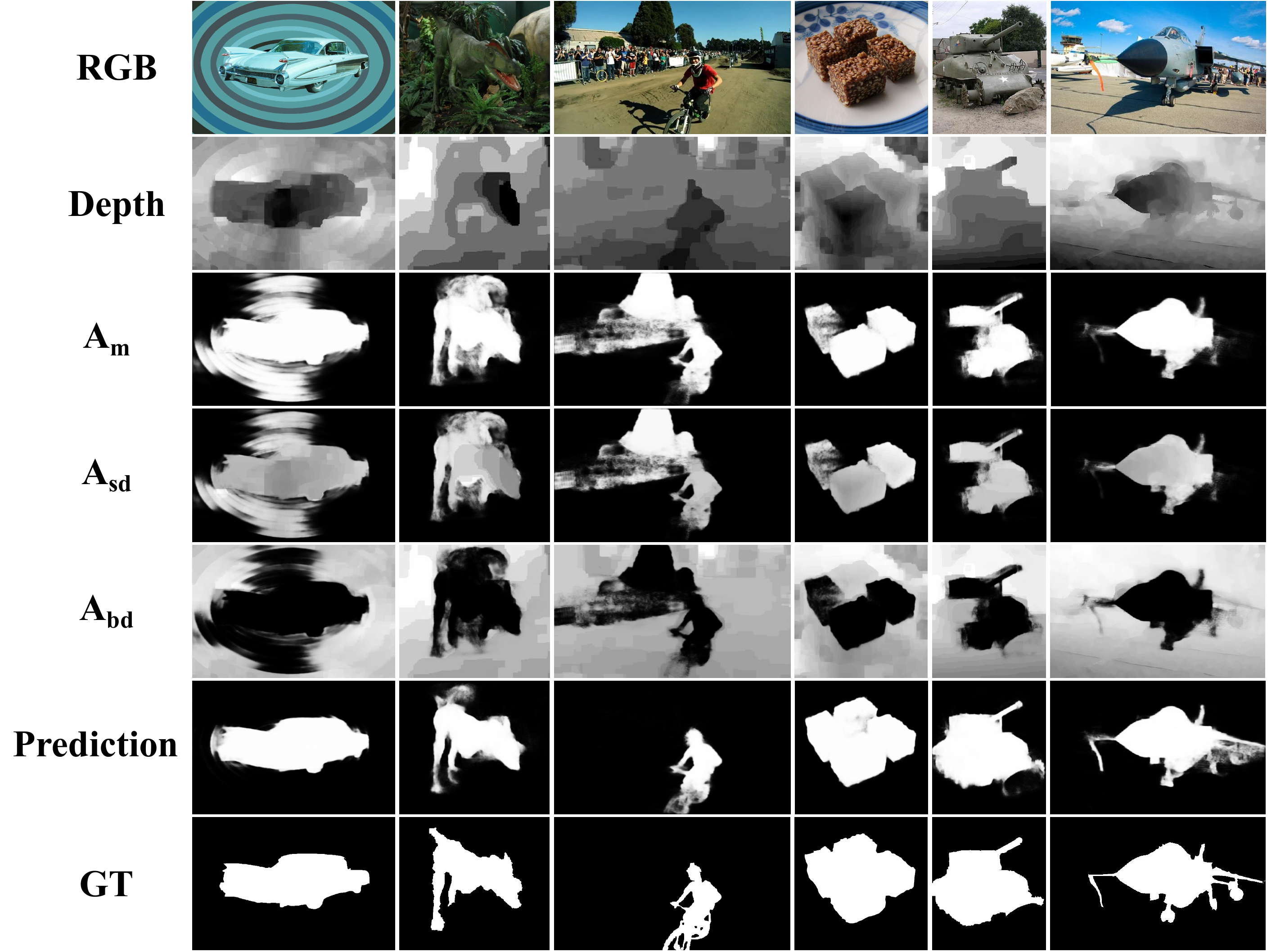}
  \caption{Visual results of using the DEDA. $A_{m}$, $A_{sd}$ and $A_{bd}$ are calculated by Equ.~\ref{equ:1}, Equ.~\ref{equ:2} and Equ.~\ref{equ:3}, respectively. }
  \label{fig:DAM_visual}
  \end{figure}

\textbf{Effectiveness of Depth-Enhanced Dual Attention Module.} 
We compare three attention modules based on the `Cat$_{He}$' model. The results are shown in Tab.~\ref{tab:ablation}. We try to directly use the depth map as the attention between the encoder and decoder. Since the depth value often varies widely inside the foreground or the background, it easily misleads salient object segmentation and performs badly, even worse than the Cat$_{He}$ model. To this end, we use the mask-guided attention (MGA) and the performance is indeed improved. Based on it, we further introduce the depth guidance and build two attended branches to form the depth-enhanced dual attention module (DEDA). It can be seen that the DEFA and DEDA achieve significant performance improvement compared to the MGA. And, the gap between the DEFA and DEDA indicates that the background branch has important supplement to the final prediction. I should note is that we do not deeply consider the two-branch fusion. Since the output of each branch is only a single-channel map, it might not produce too much performance improvement no matter what fusion is used.  
In addition, we qualitatively show the benefits of the DEDA in Fig~\ref{fig:DAM_visual}. It can be seen that the mask-guided attention wrongly classifies some salient regions as the background (see the $1^{st}$ - $3^{th}$ columns) and some background regions to be salient (see the $4^{th}$ - $6^{th}$ columns). By introducing extra contrast cues provided by the depth map for these regions, the decoder can very well correct some mistakes in the final predictions.
  
\textbf{Effectiveness of Pyramidally Attended Feature Extraction.} To be fair, we compare the PAFE with the ASPP which also uses the same convolution operations. That is, both the two modules equip a $1 \times 1$ convolution, three $3 \times 3$ atrous convolution with dilation rates of $[2, 4, 6]$ and a global average pooling. The results in Tab.~\ref{tab:ablation} indicate that the proposed PAFE is more competitive than the ASPP. In addition, we also compare them in terms of Flops ($4.00$G vs. $3.86$G) and Params ($7.07$M vs. $6.82$M). Our PAFE does not increase much more computation cost.
\section{Conclusions}
In this paper, a more efficient way of using depth information is proposed. We build a single-stream network with the novel depth-enhanced dual attention for real-time and robust salient object detection. We first abandon the routines of the two-stream cross-modal fusion and design a single stream encoder to make full use of the representation ability of the pre-trained network. Next, we use the depth-enhanced dual attention module to make the decoder jointly optimize the fore-/back-ground predictions. Benefiting from the above two ingenious designs, the saliency detection performance is greatly improved while almost no parameters are increased. In addition, we introduce the self-attention mechanism to pyramidally weight multi-scale features, thereby obtaining accurate contextual information to guide salient object segmentation.
Extensive experimental results demonstrate that the proposed model notably outperforms ten state-of-the-art methods under different evaluation metrics. Moreover, our model size is only 106.7 MB with the VGG-16 backbone and runs a real-time speed of 32 FPS. 
\\  
\\
\textbf{Acknowledgements.}
This work was supported in part by the
National Key R\&D Program of China \#2018AAA0102003,
National Natural Science Foundation of China \#61876202, \#61725202, \#61751212 and \#61829102,
the Dalian Science and Technology Innovation Foundation \#2019J12GX039, 
and the Fundamental Research Funds for the Central Universities \# DUT20ZD212.  

\clearpage
% ---- Bibliography ----
%
% BibTeX users should specify bibliography style 'splncs04'.
% References will then be sorted and formatted in the correct style.
%
\bibliographystyle{splncs04}
\bibliography{egbib}

\begin{thebibliography}{10}
\providecommand{\url}[1]{\texttt{#1}}
\providecommand{\urlprefix}{URL }
\providecommand{\doi}[1]{https://doi.org/#1}

\bibitem{colorcontrast_Fm}
Achanta, R., Hemami, S., Estrada, F., S{\"u}sstrunk, S.: Frequency-tuned
  salient region detection. In: CVPR. pp. 1597--1604 (2009)

\bibitem{PCA}
Chen, H., Li, Y.: Progressively complementarity-aware fusion network for rgb-d
  salient object detection. In: CVPR. pp. 3051--3060 (2018)

\bibitem{TANet}
Chen, H., Li, Y.: Three-stream attention-aware network for rgb-d salient object
  detection. IEEE TIP  \textbf{28}(6),  2825--2835 (2019)

\bibitem{MMCI}
Chen, H., Li, Y., Su, D.: Multi-modal fusion network with multi-scale
  multi-path and cross-modal interactions for rgb-d salient object detection.
  Pattern Recognition  \textbf{86},  376--385 (2019)

\bibitem{Deeplab}
Chen, L.C., Papandreou, G., Kokkinos, I., Murphy, K., Yuille, A.L.: Deeplab:
  Semantic image segmentation with deep convolutional nets, atrous convolution,
  and fully connected crfs. IEEE TPAMI  \textbf{40}(4),  834--848 (2017)

\bibitem{RAS}
Chen, S., Tan, X., Wang, B., Hu, X.: Reverse attention for salient object
  detection. In: ECCV. pp. 234--250 (2018)

\bibitem{RGBD135}
Cheng, Y., Fu, H., Wei, X., Xiao, J., Cao, X.: Depth enhanced saliency
  detection method. In: International Conference on Internet Multimedia
  Computing and Service. p.~23 (2014)

\bibitem{DCMC}
Cong, R., Lei, J., Zhang, C., Huang, Q., Cao, X., Hou, C.: Saliency detection
  for stereoscopic images based on depth confidence analysis and multiple cues
  fusion. IEEE SPL  \textbf{23}(6),  819--823 (2016)

\bibitem{R3Net}
Deng, Z., Hu, X., Zhu, L., Xu, X., Qin, J., Han, G., Heng, P.A.: R3net:
  Recurrent residual refinement network for saliency detection. In: IJCAI. pp.
  684--690 (2018)

\bibitem{S-m}
Fan, D.P., Cheng, M.M., Liu, Y., Li, T., Borji, A.: Structure-measure: A new
  way to evaluate foreground maps. In: ICCV. pp. 4548--4557 (2017)

\bibitem{Em}
Fan, D.P., Gong, C., Cao, Y., Ren, B., Cheng, M.M., Borji, A.:
  Enhanced-alignment measure for binary foreground map evaluation. arXiv
  preprint arXiv:1805.10421  (2018)

\bibitem{SIP}
Fan, D.P., Lin, Z., Zhao, J.X., Liu, Y., Zhang, Z., Hou, Q., Zhu, M., Cheng,
  M.M.: Rethinking rgb-d salient object detection: Models, datasets, and
  large-scale benchmarks. arXiv preprint arXiv:1907.06781  (2019)

\bibitem{late_fusion}
Fan, X., Liu, Z., Sun, G.: Salient region detection for stereoscopic images.
  In: International Conference on Digital Signal Processing. pp. 454--458
  (2014)

\bibitem{Imagecaption}
Fang, H., Gupta, S., Iandola, F., Srivastava, R.K., Deng, L., Doll{\'a}r, P.,
  Gao, J., He, X., Mitchell, M., Platt, J.C., et~al.: From captions to visual
  concepts and back. In: CVPR. pp. 1473--1482 (2015)

\bibitem{Middle_fusion}
Feng, D., Barnes, N., You, S., McCarthy, C.: Local background enclosure for
  rgb-d salient object detection. In: CVPR. pp. 2343--2350 (2016)

\bibitem{CTMF}
Han, J., Chen, H., Liu, N., Yan, C., Li, X.: Cnns-based rgb-d saliency
  detection via cross-view transfer and multiview fusion. IEEE Transactions on
  Cybernetics  \textbf{48}(11),  3171--3183 (2017)

\bibitem{PRelu}
He, K., Zhang, X., Ren, S., Sun, J.: Delving deep into rectifiers: Surpassing
  human-level performance on imagenet classification. In: ICCV. pp. 1026--1034
  (2015)

\bibitem{NJU2000}
Ju, R., Ge, L., Geng, W., Ren, T., Wu, G.: Depth saliency based on anisotropic
  center-surround difference. In: ICIP. pp. 1115--1119 (2014)

\bibitem{FPN}
Lin, T.Y., Doll{\'a}r, P., Girshick, R., He, K., Hariharan, B., Belongie, S.:
  Feature pyramid networks for object detection. In: CVPR. pp. 2117--2125
  (2017)

\bibitem{poly}
Liu, W., Rabinovich, A., Berg, A.C.: Parsenet: Looking wider to see better.
  arXiv preprint arXiv:1506.04579  (2015)

\bibitem{tracking}
Mahadevan, V., Vasconcelos, N.: Saliency-based discriminant tracking. In: CVPR
  (2009)

\bibitem{Fwb}
Margolin, R., Zelnik-Manor, L., Tal, A.: How to evaluate foreground maps? In:
  CVPR. pp. 248--255 (2014)

\bibitem{HDFNet}
Pang, Y., Zhang, L., Zhao, X., Lu, H.: Hierarchical dynamic filtering network
  for rgb-d salient object detection. In: ECCV (2020)

\bibitem{MINet}
Pang, Y., Zhao, X., Zhang, L., Lu, H.: Multi-scale interactive network for
  salient object detection. In: CVPR. pp. 9413--9422 (2020)

\bibitem{early_fusion_1}
Peng, H., Li, B., Xiong, W., Hu, W., Ji, R.: Rgbd salient object detection: A
  benchmark and algorithms. In: ECCV. pp. 92--109 (2014)

\bibitem{DMRA}
Piao, Y., Ji, W., Li, J., Zhang, M., Lu, H.: Depth-induced multi-scale
  recurrent attention network for saliency detection. In: ICCV. pp. 7254--7263
  (2019)

\bibitem{BASNet}
Qin, X., Zhang, Z., Huang, C., Gao, C., Dehghan, M., Jagersand, M.: Basnet:
  Boundary-aware salient object detection. In: CVPR. pp. 7479--7489 (2019)

\bibitem{DF}
Qu, L., He, S., Zhang, J., Tian, J., Tang, Y., Yang, Q.: Rgbd salient object
  detection via deep fusion. IEEE TIP  \textbf{26}(5),  2274--2285 (2017)

\bibitem{classification}
Ren, Z., Gao, S., Chia, L.T., Tsang, I.W.H.: Region-based saliency detection
  and its application in object recognition. IEEE TCSVT  \textbf{24}(5),
  769--779 (2013)

\bibitem{Reid}
Rui, Z., Ouyang, W., Wang, X.: Unsupervised salience learning for person
  re-identification. In: CVPR (2013)

\bibitem{VGG}
Simonyan, K., Zisserman, A.: Very deep convolutional networks for large-scale
  image recognition. arXiv preprint arXiv:1409.1556  (2014)

\bibitem{early_fusion_2}
Song, H., Liu, Z., Du, H., Sun, G., Le~Meur, O., Ren, T.: Depth-aware salient
  object detection and segmentation via multiscale discriminative saliency
  fusion and bootstrap learning. IEEE TIP  \textbf{26}(9),  4204--4216 (2017)

\bibitem{RFCN}
Wang, L., Wang, L., Lu, H., Zhang, P., Ruan, X.: Saliency detection with
  recurrent fully convolutional networks. In: ECCV. pp. 825--841 (2016)

\bibitem{AF_RGBD}
Wang, N., Gong, X.: Adaptive fusion for rgb-d salient object detection. IEEE
  Access  \textbf{7},  55277--55284 (2019)

\bibitem{SRM}
Wang, T., Borji, A., Zhang, L., Zhang, P., Lu, H.: A stagewise refinement model
  for detecting salient objects in images. In: ICCV. pp. 4019--4028 (2017)

\bibitem{LFSD_CNNs}
Wang, T., Piao, Y., Li, X., Zhang, L., Lu, H.: Deep learning for light field
  saliency detection. In: ICCV. pp. 8838--8848 (2019)

\bibitem{DGRL}
Wang, T., Zhang, L., Wang, S., Lu, H., Yang, G., Ruan, X., Borji, A.: Detect
  globally, refine locally: A novel approach to saliency detection. In: CVPR.
  pp. 3127--3135 (2018)

\bibitem{CTBIN}
Wang, W., Shen, J., Cheng, M.M., Shao, L.: An iterative and cooperative
  top-down and bottom-up inference network for salient object detection. In:
  CVPR. pp. 5968--5977 (2019)

\bibitem{PASE}
Wang, W., Zhao, S., Shen, J., Hoi, S.C., Borji, A.: Salient object detection
  with pyramid attention and salient edges. In: CVPR. pp. 1448--1457 (2019)

\bibitem{Nonlocal}
Wang, X., Girshick, R., Gupta, A., He, K.: Non-local neural networks. In: CVPR.
  pp. 7794--7803 (2018)

\bibitem{weakly_supervised_semantic}
Wei, Y., Liang, X., Chen, Y., Shen, X., Cheng, M.M., Feng, J., Zhao, Y., Yan,
  S.: Stc: A simple to complex framework for weakly-supervised semantic
  segmentation. IEEE TPAMI  \textbf{39}(11),  2314--2320 (2016)

\bibitem{HRS}
Zeng, Y., Zhang, P., Zhang, J., Lin, Z., Lu, H.: Towards high-resolution
  salient object detection. arXiv preprint arXiv:1908.07274  (2019)

\bibitem{BMPM}
Zhang, L., Dai, J., Lu, H., He, Y., Wang, G.: A bi-directional message passing
  model for salient object detection. In: CVPR. pp. 1741--1750 (2018)

\bibitem{Amulet}
Zhang, P., Wang, D., Lu, H., Wang, H., Ruan, X.: Amulet: Aggregating
  multi-level convolutional features for salient object detection. In: ICCV.
  pp. 202--211 (2017)

\bibitem{PAGRN}
Zhang, X., Wang, T., Qi, J., Lu, H., Wang, G.: Progressive attention guided
  recurrent network for salient object detection. In: CVPR. pp. 714--722 (2018)

\bibitem{CPFP}
Zhao, J.X., Cao, Y., Fan, D.P., Cheng, M.M., Li, X.Y., Zhang, L.: Contrast
  prior and fluid pyramid integration for rgbd salient object detection. In:
  CVPR (2019)

\bibitem{PFA}
Zhao, T., Wu, X.: Pyramid feature attention network for saliency detection. In:
  CVPR. pp. 3085--3094 (2019)

\bibitem{GateNet}
Zhao, X., Pang, Y., Zhang, L., Lu, H., Zhang, L.: Suppress and balance: A
  simple gated network for salient object detection. In: ECCV (2020)

\bibitem{PDNet}
Zhu, C., Cai, X., Huang, K., Li, T.H., Li, G.: Pdnet: Prior-model guided
  depth-enhanced network for salient object detection. In: ICME. pp. 199--204
  (2019)

\bibitem{SSD}
Zhu, C., Li, G.: A three-pathway psychobiological framework of salient object
  detection using stereoscopic technology. In: ICCV. pp. 3008--3014 (2017)

\bibitem{CDCP}
Zhu, C., Li, G., Wang, W., Wang, R.: An innovative salient object detection
  using center-dark channel prior. In: ICCV. pp. 1509--1515 (2017)

\bibitem{image_editing}
Zhu, J.Y., Wu, J., Xu, Y., Chang, E., Tu, Z.: Unsupervised object class
  discovery via saliency-guided multiple class learning. IEEE TPAMI
  \textbf{37}(4),  862--875 (2014)

\end{thebibliography}
\end{document}